\documentclass{article}

\PassOptionsToPackage{numbers, compress}{natbib}
\usepackage[preprint]{neurips_2026}

\usepackage[utf8]{inputenc} 
\usepackage[T1]{fontenc}    
\usepackage{hyperref}       
\usepackage{url}            
\usepackage{booktabs}       
\usepackage{amsfonts}       
\usepackage{nicefrac}       
\usepackage{microtype}      
\usepackage{xcolor}         
\usepackage{graphicx}       
\usepackage{amsmath}
\usepackage{multirow}
\usepackage{enumitem}
\usepackage{caption}
\usepackage{subcaption}
\usepackage{algorithm}
\usepackage{algorithmic}
\usepackage{amssymb}

\title{
CAPE: Contrastive Action-conditioned Parallel Encoding for Embodied Planning
}

\author{%
   Cong Chen \\
  \And
  Haowen Wang \\
  \And
  Zhixiang Zhang \\
  \And
  Pei Ren \\
  \And
  Zhengping Che \\
}

\begin{document}

\maketitle

\begin{abstract}

Embodied agents need to predict the future consequences of candidate actions in order to plan effectively before execution. 
Existing visual dynamics models learn by reconstructing future visual states or rolling out dense latent representations, 
which spreads learning capacity across visually salient but planning-irrelevant content rather than the action-conditioned changes that drive manipulation outcomes. 
We propose CAPE, a Contrastive Action-conditioned Parallel Encoding framework that learns visual dynamics by distinguishing the future outcomes induced by different action sequences. 
Given an initial observation and a candidate action sequence,
CAPE decodes the full future latent trajectory in a single forward pass 
and is trained with a Goal-Convergent Contrastive Objective that aligns predictions corresponding to the same future outcome while separating those corresponding to different outcomes. 
On real-world DROID and zero-shot transfer to RoboCasa, 
CAPE substantially outperforms prior baselines on future-state retrieval, offline action matching, and closed-loop planning, while notably reducing planning-time inference cost at long prediction horizons.

\end{abstract}

\section{Introduction}

The ability to predict how the environment will evolve under candidate actions is fundamental to planning for embodied agents in robotic manipulation~\citep{williams2017information, guo2026flowdreamer, hafner2023dreamerv3, zhou2024dinowm, fragkiadaki2016learning}.
In vision-based settings, this evolution is manifested as the visual changes induced by candidate actions~\citep{finn2016unsupervised, oh2015action, micheli2023IRIS}.
Building on this view, recent work has increasingly focused on learning action-conditioned forward dynamics models that predict such visual transitions, 
providing the backbone of modern model-based embodied planning~\citep{finn2017deep, ebert2018visual, dasari2019robonet, ebert2017self}.

\begin{figure}[tbp]
  \centering
  \begin{subfigure}[b]{0.53\textwidth}
    \centering
    \includegraphics[width=\linewidth]{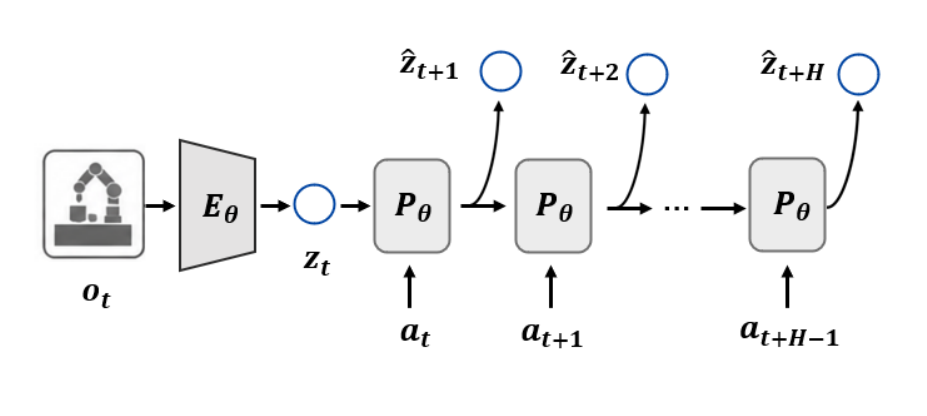}
    \caption{Autoregressive rollout}
    \label{fig:plan_ar}
  \end{subfigure}
  \hfill
  \begin{subfigure}[b]{0.45\textwidth}
    \centering
    \includegraphics[width=\linewidth]{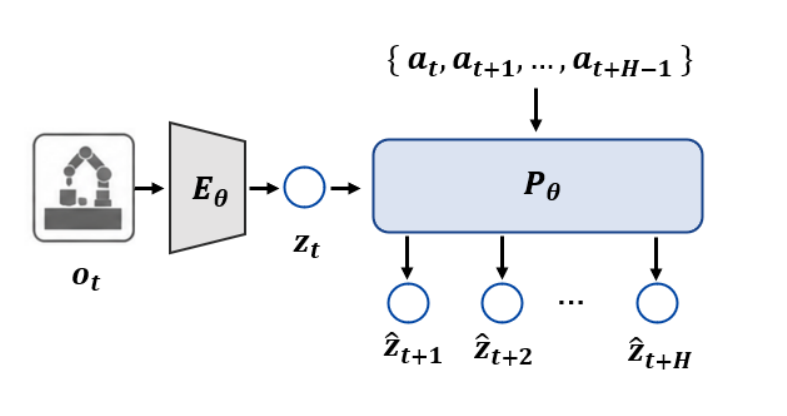}
    \caption{\textbf{Ours:} Parallel seq-to-seq prediction}
    \label{fig:plan_ours}
  \end{subfigure}
  \caption{\textbf{Comparison of different future prediction paradigms.} 
    (a) Existing latent world models unroll autoregressively, 
        requiring multiple sequential forward passes through $P_\theta$. 
    (b) Our model predicts the entire future latent trajectory 
        $(\hat{z}_{t+1}, \dots, \hat{z}_T)$ in a \textbf{single forward pass} 
        conditioned on the action sequence and current state $z_t$.}
  \label{fig:planning}
  \vspace{-5mm} 
\end{figure}

Existing approaches to action-conditioned forward dynamics modeling fall broadly into two paradigms~\citep{ai2025review, liu2025embodied, survey2025worldmodels}.
Reconstruction-based approaches model forward dynamics through pixel-level prediction from raw visual observations~\citep{ferraro2025focus, micheli2023IRIS,hafner2023dreamerv3}.
These methods have achieved strong performance in simulated reinforcement-learning environments with relatively simple backgrounds and limited visual variation.
However, in real-world manipulation scenes, pixel-level reconstruction can force the model to encode task-irrelevant appearance factors~\citep{zhang2025objects}, such as shadows, textures, and background clutter, thereby obscuring the underlying action-induced dynamics.
Latent predictive approaches instead model forward dynamics directly in the latent spaces of pre-trained visual encoders such as DINOv2~\citep{zhou2024dinowm,assran2025vjepa2,terver2025jepa-wm}.
Representative methods such as DINO-WM show that future patch tokens can be predicted directly in latent space without reconstructing pixels~\citep{zhou2024dinowm}.
However, as illustrated in Figure~\ref{fig:plan_ar}, most of these methods still rely on step-by-step rollout, requiring the model to generate a full high-dimensional future representation of visual state at every step and recursively use it to predict subsequent states.
Supervision defined over the full visual state encourages the model to reproduce large portions of the scene whose evolution is only weakly related to the agent’s actions. 
By contrast, the action-conditioned changes most relevant to planning are usually confined to a small spatial neighborhood around the end-effector and the manipulated objects, and thus contribute only weakly to the overall training objective.
This imbalance leaves the action-conditioned signal weakly represented in the learned features, limiting their reliability for downstream planning.

Our key insight is that, 
in embodied planning, action-conditioned future prediction should not be cast as faithful reconstruction of future visual states, 
but as distinguishing the future outcomes induced by different action sequences. 
This shifts the supervisory target away from the full visual state and toward the action-conditioned differences that distinguish future outcomes. 
We realize this principle through a contrastive objective over predicted future-state representations, 
aligning predictions that correspond to the same future outcome while separating those corresponding to different outcomes. 
Because much of the visual context in manipulation remains stable over a trajectory, 
a single initial observation often provides sufficient context for future prediction. 
This allows the full future trajectory to be decoded in parallel from the action sequence (Figure~\ref{fig:plan_ours}), removing the need for autoregressive rollout.

In this work, 
we introduce CAPE, a \textbf{C}ontrastive \textbf{A}ction-conditioned \textbf{P}arallel \textbf{E}ncoding framework for visual dynamics prediction.
Given an initial observation and an action sequence, 
CAPE predicts the future latent trajectory in a single forward pass, using the initial observation as shared visual context. 
It is trained with a Goal-Convergent Contrastive Objective that aligns representations corresponding to the same future outcome. 
This encourages the latent space to be organized around action-conditioned transitions rather than similarity to full future observations. 
We evaluate CAPE on the real-world DROID~\citep{khazatsky2024droid} dataset and on zero-shot transfer to customized manipulation tasks in RoboCasa~\citep{nasiriany2024robocasa}. 
CAPE substantially outperforms prior baselines on future-state retrieval, offline action matching, and closed-loop planning, while remaining significantly more computationally efficient.

Our contributions can be summarized as follows:
\begin{itemize}
[itemsep=1pt,topsep=1pt,parsep=1pt,leftmargin=20pt]

\item 
We propose CAPE, a contrastive action-conditioned framework for parallel visual dynamics prediction, which predicts future latent trajectories from a single visual context without autoregressive rollout.

\item 
We introduce a Goal-Convergent Contrastive Objective that aligns predictions corresponding to the same future outcome while separating those induced by different action sequences, 
thereby organizing the latent space around action-conditioned transitions rather than future-state appearance similarity.

\item 
We validate CAPE on both real-world and simulated benchmarks, 
demonstrating leading performance across multiple retrieval and planning evaluations together with substantially improved inference efficiency.
\end{itemize}

\section{Related Work}

\paragraph{Action-conditioned visual dynamics models.}
Learning predictive models of environment dynamics is a long-standing objective in model-based reinforcement learning and robotic control~\citep{ha2018world, kaiser2020model}, where the model is used to anticipate the consequences of candidate actions before execution. 
Early methods studied action-conditioned video prediction in Atari and robotic manipulation, learning to generate future frames from past observations and actions~\citep{oh2015action, finn2016unsupervised}.
These models were later combined with model-predictive control~\citep{garcia1989mpc} for visual planning in robotic manipulation~\citep{finn2017deep, ebert2018visual}, and large-scale robot datasets such as RoboNet~\citep{dasari2019robonet} improved generalization across robots, objects, and scenes.
Beyond pixel-space prediction, latent world models such as PlaNet~\citep{hafner2019planet}, Dreamer~\citep{hafner2020dreamer}, DreamerV2~\citep{hafner2021dreamerv2}, and DreamerV3~\citep{hafner2023dreamerv3} learn compact dynamics from visual inputs and use imagined rollouts for planning or policy learning.
More recent work further replaces pixel reconstruction with prediction in pretrained visual feature spaces: DINO-WM~\citep{zhou2024dinowm} predicts future DINOv2~\citep{oquab2023dinov2} features for zero-shot visual planning, while V-JEPA-style world models~\citep{bardes2024vjepa, assran2025vjepa2} learn predictive joint embeddings and action-conditioned latent dynamics for image-goal planning.
These approaches demonstrate the value of visual predictive models at different representation levels~\citep{bruce2024genie, seo2023masked, zhang2023storm}, but typically evaluate future consequences through sequential rollout of dense future states or features.
CAPE instead predicts the full future latent trajectory in parallel from the initial visual context and action sequence, emphasizing action-induced transitions rather than autoregressive dense-state prediction.

\paragraph{Contrastive predictive representation learning.}
Contrastive learning provides an alternative to reconstruction-based supervision by learning representations that distinguish positive pairs from negative samples.
Contrastive Predictive Coding learns temporally predictive features by predicting future latent variables with an InfoNCE objective~\citep{oord2018cpc}, while visual contrastive methods such as MoCo~\citep{he2020moco} and SimCLR~\citep{chen2020simclr} show that discriminative self-supervision can produce strong image representations without pixel-level reconstruction.
Related joint-embedding and non-contrastive approaches, including BYOL~\citep{grill2020byol}, VICReg~\citep{bardes2022vicreg}, I-JEPA~\citep{assran2023ijepa}, and V-JEPA~\citep{bardes2024vjepa}, further show that predictive representation learning can capture semantic structure while avoiding direct image generation.
In reinforcement learning and control, auxiliary contrastive representation objectives have been used to improve state abstraction and sample efficiency by encouraging features to capture temporally predictive or controllable factors~\citep{kipf2020contrastive, anand2019unsupervised, laskin2020curl,
  schwarzer2021spr, zheng2023taco, zheng2024premiertaco}.
For visual dynamics learning, this perspective is useful because supervision can be defined by whether a predicted future is consistent with the correct temporal context, action sequence, or outcome, rather than by full visual reconstruction.
CAPE adopts a contrastive objective for action-conditioned future prediction, aligning predictions that converge to the same future state while separating mismatched action-conditioned transitions.

\section{Methodology}

\subsection{Preliminaries}
Our setting is vision-based goal-conditioned planning from image observations. 
At each time step, an agent receives an image observation $o_t \in \mathbb{R}^{H \times W \times 3}$ and executes a continuous action $a_t \in \mathbb{R}^{d_a}$, 
where $d_a$ denotes the dimension of the action space. 
Given a current observation $o_t$ and a goal image $o_g$, 
the objective is to produce an action sequence $a_{t:t+H-1} = (a_t, \ldots, a_{t+H-1})$ that drives the environment toward the state depicted in $o_g$. 
Training assumes access to a pre-collected offline dataset of robot trajectories without task rewards or task labels.

Following recent latent world-model approaches~\citep{zhou2024dinowm, terver2025jepa-wm}, we formulate the problem as planning in the latent space of a pre-trained visual encoder. 
An observation encoder $enc_\phi$ maps each image to a latent representation $z_t = enc_\phi(o_t)$, 
and a one-step transition model is unrolled autoregressively to obtain multi-step futures: 
$\hat{z}_{t+1} = f_\theta(z_t, a_t)$. 
By contrast, CAPE uses the encoder output as a dense visual context $v_t = enc_\phi(o_t)$ and defines $f_\theta$ as a sequence-level mapping
\begin{equation}
\hat{z}_{t+1:t+H} = f_\theta(v_t, a_{t:t+H-1}),
\label{eq:cape-transition}
\end{equation}
where $\hat{z}_{t+h} \in \mathbb{R}^d$ denotes the predicted future-state representation at step $t+h$, 
and the full sequence is decoded in a single forward pass.

The following subsections instantiate this formulation.
Sec.~\ref{sec:framework} presents the CAPE training framework, including goal-convergent pair sampling, the parallel action-query decoder, and goal-convergent contrastive objective.
Sec.~\ref{sec:planning} then describes how the trained CAPE model is used as a latent forward model for MPC-based goal-conditioned planning.

\subsection{Contrastive Action-conditioned Parallel Encoding Framework}
\label{sec:framework}

\begin{figure*}[t]
  \centering
  \includegraphics[width=1.00\textwidth]{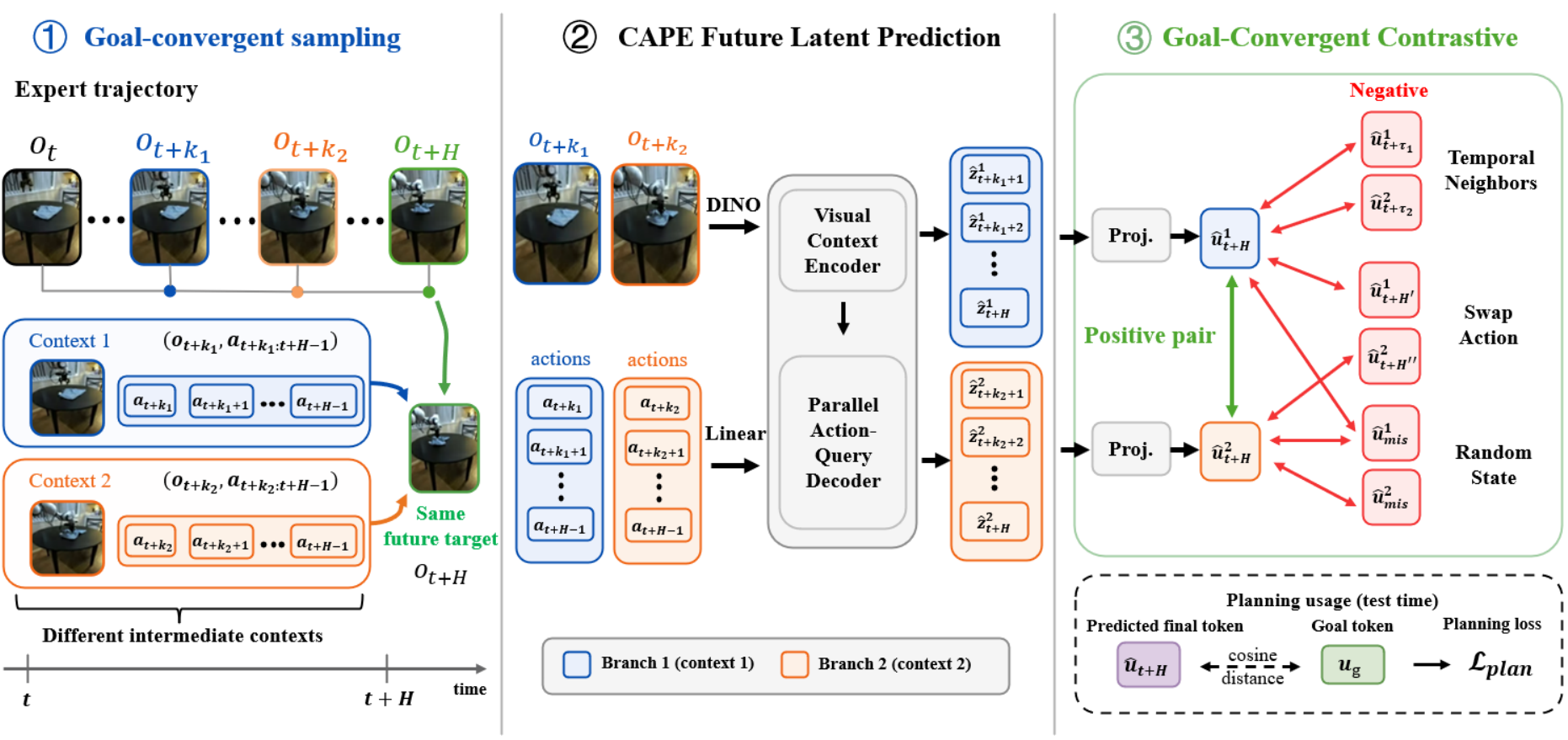}
  \caption{\textbf{Overview of CAPE framework.} 
  CAPE is trained through a three-stage goal-convergent pipeline.
    (1) Two intermediate contexts sampled from the same trajectory share the same future target at time $t+H$. 
    (2) Given each context, CAPE encodes the visual observation into context tokens and maps the corresponding action subsequence into action queries. 
    A parallel action-query decoder attends these queries to the visual context tokens and predicts future latent tokens in parallel.
    (3) The proposed goal-convergent objective aligns predictions that correspond to the same future target while repelling temporally or behaviorally mismatched predictions. 
    At test time, the predicted final token is compared with the goal token to compute the planning cost.
  }
  \label{fig:arch}
\end{figure*}

\paragraph{Parallel Action-Query Decoding Architecture.}
Given an observation $o_t$ and a candidate action sequence $a_{t:t+H-1}$, CAPE predicts the future latent trajectory $\hat{z}_{t+1:t+H}$ in a single forward pass. 
We first encode the visual observation with a frozen DINOv2~\citep{oquab2023dinov2} encoder, and obtain the current visual tokens $v_t=enc_{dino}(o_t)$. 
Since these tokens are produced by a generic visual representation model, we further process them with a lightweight visual context encoder $enc_{ctx}$, implemented as three self-attention layers: $c_t = enc_{ctx}(v_t)$.
The resulting context tokens $c_t$ provide a shared visual context that captures the scene layout, object appearance, and the local configuration from which the future transition starts. 

Rather than recursively generating the next latent state and feeding it back to predict later states, CAPE maps the action sequence into a set of horizon-indexed action queries. 
Each action $a_{t+i}$ is first mapped by a linear projection and then combined with a temporal positional embedding, producing queries 
$q_{t+1:t+H}$. 
These queries are passed to a forward dynamics model $F_{\theta}$, 
which we implement as a parallel action-query decoder. 
Specifically, $F_\theta$ consists of stacked causal self-attention and cross-attention blocks.
The causal self-attention allows each horizon-indexed action query to aggregate information from the corresponding action prefix, while the cross-attention uses the visual context tokens $c_t$ as keys and values. 
The model therefore predicts all future latent tokens in parallel:
\begin{equation}
\hat{z}_{t+1:t+H}
=
F_{\theta}\bigl(q_{t+1:t+H}, c_t\bigr).
\end{equation}

This architecture removes the need for autoregressive rollout in latent space. 
In step-wise prediction, early prediction errors are recursively consumed by subsequent steps, making long-horizon planning vulnerable to compounding latent drift.
CAPE instead uses the initial observation as a stable visual anchor and uses action queries to attend to the visual context tokens, producing future latent outcomes at all horizons in parallel.
This formulation is well suited to manipulation, where most of the scene remains static and planning-relevant changes are localized around the end-effector and manipulated objects.
Thus, rather than repeatedly reconstructing full high-dimensional visual states, CAPE learns a compact action-conditioned mapping from the current visual context and candidate actions to the future latent outcomes needed for planning.

\paragraph{Goal-Convergent Contrastive Objective.}
Direct supervision on full future visual latents can still be dominated by static or weakly action-dependent scene content. 
To bias the representation toward action-conditioned outcome differences, we train CAPE with a goal-convergent contrastive objective. 
For each expert trajectory clip, we sample two different intermediate contexts that lead to the same future target observation $o_{t+H}$. 
Specifically, we choose two offsets $k_1$ and $k_2$ and construct two branches,
$x^{1}=(o_{t+k_1}, a_{t+k_1:t+H-1}), $
$x^{2}=(o_{t+k_2}, a_{t+k_2:t+H-1}),$
which contain different visual contexts and different remaining action sequences, but converge to the same terminal outcome $o_{t+H}$. 
Passing the two branches through 
the parallel action-query decoder 
gives two predicted final latent tokens,  
$\hat{z}^{1}_{t+H}
=
F_{\theta}(q^{1}_{t+k_1+1:t+H}, c_{t+k_1}),$
$\hat{z}^{2}_{t+H}
=
F_{\theta}(q^{2}_{t+k_2+1:t+H}, c_{t+k_2}).$
Since both predictions correspond to the same future outcome, they are treated as a positive pair.

We apply a contrastive loss in the projected latent space. 
Let $g_{\phi}(\cdot)$ be an MLP projection head followed by normalization, and define
$u^{1}_{t+H}=g_{\phi}(\hat{z}^{(1)}_{t+H}),$
$u^{2}_{t+H}=g_{\phi}(\hat{z}^{(2)}_{t+H}).$
For notational simplicity, we denote the projected final-horizon representations as $u^{1}$ and $u^{2}$ inside the loss. 
With cosine similarity $s(\cdot,\cdot)$ and temperature $\tau$, the goal-convergent contrastive loss is
\begin{equation}
\mathcal{L}_{\mathrm{gc}}
=
-\log
\frac{
\exp(s(u^{(1)},u^{(2)})/\tau)
}{
\exp(s(u^{(1)},u^{(2)})/\tau)
+
\sum_{v \in \mathcal{N}}
\exp(s(u^{(1)},v)/\tau)
},
\end{equation}
with a symmetric term obtained by swapping the two branches. 

The negative set $\mathcal{N}$ contains states that should be distinguished from the target goal, including temporal neighbors from nearby but distinct future time steps, outcomes produced by swapped or mismatched action sequences, and randomly sampled states from other trajectories. 
Temporal neighbors make the discrimination fine-grained, since they often share most visual content with the target but differ in the precise action-induced outcome. 
Swapped-action negatives prevent the model from collapsing onto visual context alone by requiring the representation to distinguish different futures reachable from similar observations. 
Random-state negatives provide broader separation across unrelated scenes and tasks.

This objective recasts future prediction supervision from full-state reconstruction to outcome discrimination.
Rather than requiring the model to faithfully reproduce every component of the future visual state, CAPE learns to align action-conditioned transitions that lead to the same future outcome.
Consequently, the learned latent space is organized around goal-convergent dynamics rather than raw visual similarity.

\subsection{Planning with CAPE via Model Predictive Control}
\label{sec:planning}

We embed the trained CAPE model as the dynamics model within a Model Predictive Control (MPC) framework. 
At each environment step, the planner receives the current observation $o_t$ and a visual goal observation $o_g$. 
The current observation is encoded into visual context tokens $c_t$, while 
the goal observation $o_g$ is mapped through the same visual-context pathway and a lightweight goal projection module, yielding a goal token $z_g$ that is aligned with the action-conditioned prediction $\hat{z}_{t+H}$. 
The planner then optimizes over a horizon-$H$ action sequence 
$A = \{a_t,\ldots,a_{t+H-1}\}$ using the Cross-Entropy Method (CEM).

For each sampled candidate sequence $A$, CAPE predicts the future latent trajectory in a single forward pass,
$\hat{z}_{t+1:t+H} = F_\theta(q_{t+1:t+H}, c_t),$
using the parallel action-query decoder described in Section~\ref{sec:framework}. 
The candidate is evaluated by the cosine distance in the same projected outcome space used during contrastive training, with $\hat{u}_{t+H}=g_\phi(\hat{z}_{t+H})$ and $u_g=g_\phi(z_g)$.
We define 
\begin{equation}
L_{plan}(A)
=
1 -
\frac{\hat{u}_{t+H}^{\top} u_g}
{\|\hat{u}_{t+H}\|_2 \|u_g\|_2}.
\end{equation}
CEM iteratively samples action sequences, selects the elite candidates with the lowest cost, and refits the sampling distribution around them. 
After optimization, the first action of the best sequence is executed, and the procedure is repeated at the next time step.

This planning procedure directly benefits from the structure learned by the goal-convergent contrastive objective. 
Because the terminal prediction token is trained to distinguish action-conditioned outcomes rather than reconstruct full visual states, the MPC cost is less dominated by action-irrelevant appearance variation. 
Moreover, since all candidate trajectories are evaluated without autoregressive latent rollout, CAPE substantially reduces the computational cost of CEM while avoiding the accumulation of step-wise prediction errors.

\section{Experiments}

\subsection{Experimental Setup}
\label{sec:exp-setup}

\paragraph{Datasets and Tasks.}
We use DROID~\citep{khazatsky2024droid} as the primary real-world benchmark for model training, representation analysis, and offline planning evaluation. 
We design a future-state retrieval task to evaluate whether the learned representations are discriminative, 
and an offline action-matching task to evaluate the precision of downstream action planning.
To evaluate cross-domain generalization, we further use the RoboCasa~\citep{nasiriany2024robocasa} simulation environment. 
RoboCasa~\citep{nasiriany2024robocasa} provides diverse household manipulation scenes and enables controlled closed-loop evaluation under visual and task-domain shifts.
We follow the zero-shot RoboCasa~\citep{nasiriany2024robocasa} evaluation protocol of JEPA-WM~\citep{terver2025jepa-wm}, and evaluate on custom pick-and-place tasks defined from teleoperated trajectories, namely ``Place'' and ``Reach'', denoted Rc-Pl and Rc-R.

\paragraph{Evaluation Protocols.}
We primarily focus on metrics across the following three dimensions:
1) \textbf{Hit@K:} For future state retrieval task, we evaluate the performance using Hit@K as metrics, which denotes the percentage of queries whose ground-truth future state is ranked within the top-K
candidates. 
2) \textbf{Action Score:} For real-world offline action matching task on DROID~\citep{khazatsky2024droid}, we follow JEPA-WM~\citep{terver2025jepa-wm} and evaluate action matching between the planned action sequence and the ground-truth action sequence under the same initial and goal observations. 
The resulting L1 error is inverted and linearly rescaled into \textit{Action Score}, where higher is better. 
3) \textbf{Model Predictive Control Success Rate(SR $\uparrow$):} For zero-shot evaluation on RoboCasa~\citep{nasiriany2024robocasa}, we evaluate the agent's success rate in reaching the designed goal state under MPC planning.

\paragraph{Baselines.}
We compare CAPE against several representative baselines that internally model action-conditioned forward dynamics.
These methods are categorized into reconstruction-based, latent predictive, and temporal-contrastive approaches, as illustrated in Tab.~\ref{tab:retrieval}.
For reconstruction-based methods, we include IRIS~\citep{micheli2023IRIS}, which tokenizes visual states with a discrete autoencoder and predicts future tokens conditioned on action autoregressively using a Transformer, and DreamerV3~\citep{hafner2023dreamerv3}, which learns categorical latent state representations for imagination-based control.
For latent predictive methods, we compare with DINO-WM~\citep{zhou2024dinowm}, V-JEPA-2-AC~\citep{assran2025vjepa2}, and JEPA-WM~\citep{terver2025jepa-wm}, which predict future latent representations rather than reconstructing pixels or visual tokens.
For temporal-contrastive methods, we include Premier-TACO~\citep{zheng2024premiertaco}, which learns action-conditioned temporal representations through contrastive objectives for downstream policy learning.
We also include a zero-action baseline to quantify the contribution of action-conditioned dynamics. 
To ensure a fair comparison, we use the same evaluation environments and MPC planner configuration as JEPA-WM~\citep{terver2025jepa-wm}.
Additional baseline adaptation details and capacity comparisons are provided in Appendix~\ref{sec:app-b}.

\subsection{Future State Retrieval}
\label{sec:exp-retrieval}

\begin{table}[h]
  \centering
  \caption{
  Future state retrieval results on the DROID dataset. 
  Models are evaluated at three prediction horizons: $t{+}1$ (0.4s), $t{+}3$ (1.2s), and $t{+}5$ (2.0s), using Hit@1, Hit@5, and Hit@10 as metrics. 
  For Premier-TACO~\citep{zheng2024premiertaco}, we train a separate model for each horizon, as its formulation requires a fixed prediction horizon $h$.
  }
  \label{tab:retrieval} 
  \resizebox{\textwidth}{!}{%
  \begin{tabular}{@{}llccccccccc@{}}
    \toprule
    \multirow{2}{*}{\textbf{Method Category}} & \multirow{2}{*}{\textbf{Model}} & \multicolumn{3}{c}{\textbf{t+1 (0.4s)}} & \multicolumn{3}{c}{\textbf{t+3 (1.2s)}} & \multicolumn{3}{c}{\textbf{t+5 (2.0s)}} \\
    \cmidrule(lr){3-5} \cmidrule(lr){6-8} \cmidrule(lr){9-11}
    & & \textbf{Hit@1} & \textbf{Hit@5} & \textbf{Hit@10} & \textbf{Hit@1} & \textbf{Hit@5} & \textbf{Hit@10} & \textbf{Hit@1} & \textbf{Hit@5} & \textbf{Hit@10} \\
    \midrule
    \multirow{2}{*}{Reconstruction}
    & IRIS & 4.95 & 90.90 & \textbf{96.84} & 0.55 & 30.12 & 63.35 & 0.80 & 8.59 & 28.50 \\
    & DreamerV3 & 5.54 & 26.22 & 35.03 & 2.92 & 19.76 & 30.85 & 2.28 & 15.65 & 26.21 \\
    \midrule
    \multirow{3}{*}{Latent Predictive} 
    & DINO-WM & 7.66 & 70.87 & 85.73 & 4.06 & 26.21 & 48.91 & 2.86 & 15.35 & 31.89 \\
    & V-JEPA2 AC & 1.23 & 66.64 & 84.02 & 1.07 & 26.20 & 49.10 & 1.47 & 17.51 & 37.77 \\
    & JEPA-WM & 5.15 & 69.15 & 83.63 & 4.20 & 25.21 & 45.62 & 3.29 & 16.13 & 32.02 \\
    \midrule
    Temporal Contrastive & Premier-TACO & 5.14 & 28.62 & 36.06 & 3.61 & 16.40 & 27.99 & 4.36 & 14.88 & 23.14 \\
    \midrule
    Non-predictive & zero-action & 6.07 & 23.05 & 33.42 & 4.08 & 16.35 & 24.95 & 2.52 & 11.73 & 18.30 \\
    \midrule
    \textbf{Ours} & \textbf{CAPE} & \textbf{54.11} & \textbf{91.64} & 96.13 & \textbf{48.27} & \textbf{87.48} & \textbf{95.17} & \textbf{42.97} & \textbf{84.19} & \textbf{93.68} \\
    \bottomrule
  \end{tabular}%
  }
\end{table}
To evaluate action-conditioned predictive representations independently of downstream planning, we evaluate future-state retrieval on held-out DROID~\citep{khazatsky2024droid} trajectories. 
Given a current observation and a future action sequence, the model predicts a future state representation, which is then used to retrieve the correct future frame from a candidate pool. 
This task evaluates whether the learned dynamics can produce future representations that are accurate under the specified actions and discriminative among visually similar candidate observations.

Tab.~\ref{tab:retrieval} shows that CAPE achieves the strongest overall performance among the evaluated methods, with the best results on nearly all reported horizons.
We observed that, while prior works exhibit a sharp performance drop as the predictive horizon increases, which is consistent with the accumulation of rollout error over long horizons, the retrieval performance of CAPE degrades much more slowly.
Notably, CAPE achieves exceptionally high Hit@1 scores, reaching 54.11\% at $t+1$ and maintaining 42.97\% at $t+5$.
In contrast, all other predictive models remain below 8\% Hit@1 across the same horizons.
These empirical results indicate that the future state representations learned by CAPE remain substantially more discriminative as the horizon increases.

\begin{table*}[!ht]
  \centering
  \normalsize 
  \caption{\textbf{Downstream Planning Performance.} We evaluate offline action matching via Action Score ($\uparrow$) on the DROID dataset and closed-loop zero-shot planning via Success Rate (SR \%, $\uparrow$) on RoboCasa tasks (Rc-R and Rc-Pl). $h$ denotes the prediction horizon. The baselines are grouped by their underlying representation paradigms.}
  \label{tab:planning_performance}
  \resizebox{0.75\textwidth}{!}{ 
  \begin{tabular}{lcccccc}
    \toprule
    \multirow{2}{*}{\textbf{Method}} & \multicolumn{2}{c}{\textbf{DROID (Action Score $\uparrow$)}} & \multicolumn{2}{c}{\textbf{Rc-R (SR \% $\uparrow$)}} & \multicolumn{2}{c}{\textbf{Rc-Pl (SR \% $\uparrow$)}} \\
    \cmidrule(lr){2-3} \cmidrule(lr){4-5} \cmidrule(lr){6-7}
    & $h=3$ & $h=5$ & $h=3$ & $h=5$ & $h=3$ & $h=5$ \\
    \midrule
    IRIS & 22.7 & 17.7 & 9.4 & 4.7 & 2.4 & 0.0 \\
    DreamerV3 & 14.3 & 11.5 & 4.2 & 2.3 & 2.7 & 0.3 \\
    \midrule
    DINO-WM & 37.4 & 24.6 & 16.8 & 7.6 & 26.3 & 16.3 \\
    V-JEPA2 AC & 39.6 & 28.8 & 20.2 & 17.3 & 31.1 & 20.6 \\
    JEPA-WM & 44.2 & 32.8 & 23.4 & 18.4 & \textbf{34.4} & \textbf{26.7} \\
    \midrule
    Premier-TACO & 11.3 & 7.6 & 3.3 & 1.3 & 2.6 & 0.0 \\
    \midrule
    \textbf{CAPE (Ours)} & \textbf{62.1} & \textbf{56.6} & \textbf{49.0} & \textbf{32.1} & 28.9 & 22.5 \\
    \bottomrule
  \end{tabular}
  }
\end{table*}

\begin{table}[!ht]
  \centering
  \normalsize 
  \caption{\textbf{Computational Efficiency of CEM Planning.} We report the single MPC decision loop  time (in milliseconds $\downarrow$) required to evaluate action sequences of length $h$. 
  All timing evaluations are conducted on a single NVIDIA RTX 4090 GPU.}
  \label{tab:efficiency}
  \resizebox{0.6\linewidth}{!} {
  \begin{tabular}{lccccc}
    \toprule
    \multirow{2}{*}{\textbf{Method}} & \multirow{2}{*}{\textbf{Samples}} & \multirow{2}{*}{\textbf{Iter.}} & \multicolumn{3}{c}{\textbf{Total Planning Time (ms) $\downarrow$}} \\
    \cmidrule(lr){4-6}
    & & & $h=1$ & $h=3$ & $h=5$ \\
    \midrule
    IRIS & 300 & 15 & 1560 & 13354 & 43900 \\
    DreamerV3 & 300 & 15 & \textbf{26} & \textbf{64} & \textbf{105} \\
    DINO-WM & 300 & 15 & 1978 & 11396 & 20863 \\
    \midrule
    \textbf{CAPE (Ours)} & 300 & 15 & \underline{335} & \underline{340} & \underline{356} \\
    \bottomrule
  \end{tabular}
  }
\end{table}
\subsection{Planning Performance and Efficiency}
\label{sec:exp-planning}

\paragraph{Planning Performance.}
For all methods, planning is performed with the same CEM-based MPC framework, so that differences reflect the quality and efficiency of the learned world models rather than the planner itself.
Tab.~\ref{tab:planning_performance} summarizes the performance of our framework under two complementary evaluation settings: offline action matching on the real-world DROID~\citep{khazatsky2024droid} dataset, and closed-loop zero-shot planning in the RoboCasa~\citep{nasiriany2024robocasa} simulation environment (Rc-R and Rc-Pl).
Across both offline Action Score and online Success Rate metrics, CAPE demonstrates competitive planning accuracy.
On the real-world DROID~\citep{khazatsky2024droid} benchmark, it achieves the highest Action Score at both horizons, and in RoboCasa~\citep{nasiriany2024robocasa} it delivers the strongest results on Rc-R while remaining competitive on Rc-Pl.

\paragraph{Computational Efficiency.}
Tab.~\ref{tab:efficiency} reports the wall-clock time of a single CEM-based MPC decision loop.
CAPE is substantially more efficient than autoregressive baselines, while maintaining strong planning performance. 
Most notably, planning time of our CAPE remains nearly constant as the prediction horizon increases, rising only from 335ms at $h=1$ to 356ms at $h=5$. 
In contrast, autoregressive models such as IRIS~\citep{micheli2023IRIS} and DINO-WM~\citep{zhou2024dinowm} become dramatically slower with longer horizons, since each candidate trajectory requires repeated multi-step rollout. 
Although DreamerV3~\citep{hafner2023dreamerv3} is faster in absolute runtime, its planning accuracy is substantially lower than that of CAPE in our evaluations. 
This suggests that CAPE achieves a more favorable trade-off between planning performance and computational efficiency.
These results show that our parallel prediction design of CAPE is considerably better suited to sampling-based planning, where many candidate action sequences must be evaluated within each optimization step.

\begin{figure*}[!ht]
  \centering
  \includegraphics[width=1.00\textwidth]{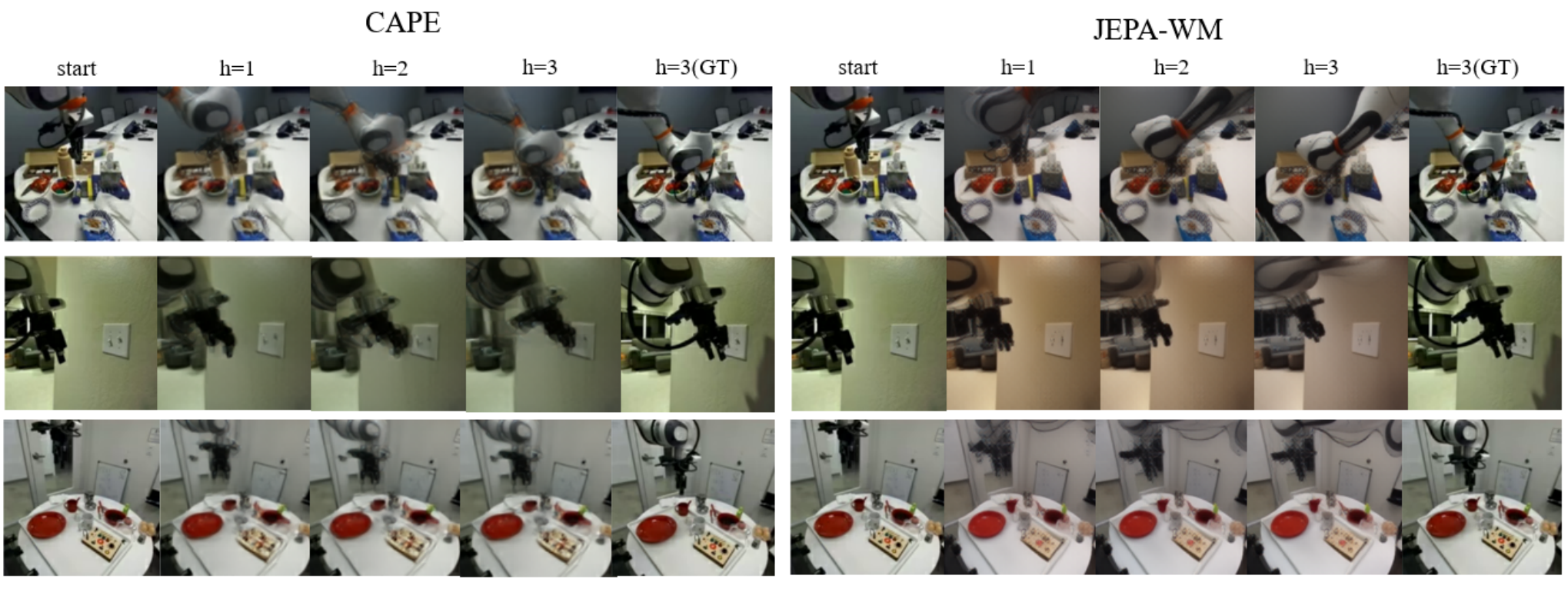}
  \caption{\textbf{Qualitative comparison of multi-step future prediction.}
  Given the same start observation, we compare the multi-step future predictions of our method, with an autoregressive baseline~\citep{terver2025jepa-wm}.
  CAPE produces more temporally coherent and goal-consistent predictions, better preserving robot configuration over longer horizons.
  }
  \label{fig:vis_rollout}
\end{figure*}

\begin{figure*}[!ht]
  \centering
  \includegraphics[width=1.00\linewidth]{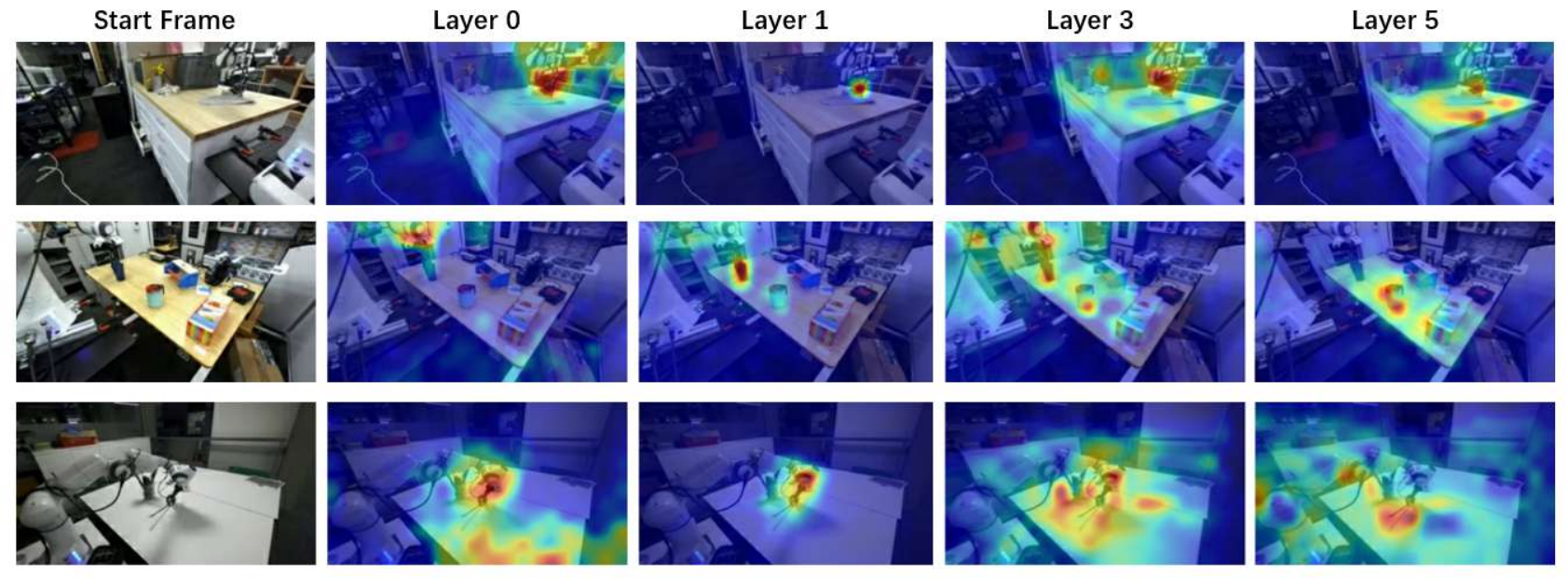}
  \caption{\textbf{Visualization of Action-query attention over visual context.}
  We visualize the cross-attention weights from the action query to the spatial visual context tokens extracted from the input start frame across network depth.
  }
  \label{fig:attn_map}
\end{figure*} 
\subsection{Visual Analysis of CAPE}

\paragraph{Qualitative comparison.} 

To illustrate that CAPE learns compact yet informative representations for action-conditioned future prediction, we visualize multi-step future predictions induced by the same initial observation and action sequence under different prediction paradigms.
Specifically, given a start frame and an action sequence of horizon three, we compare the predicted intermediate future states of CAPE with those produced by JEPA-WM, which follows an autoregressive state-by-state rollout paradigm.
As shown in Fig.~\ref{fig:vis_rollout}, even when the initial observation does not fully reveal the robot arm, CAPE still predicts future representations that capture the correct robot motion across different viewpoints.
In contrast, the autoregressive rollout of JEPA-WM incorrectly infers the configuration of the robot arm outside the visible field of view, leading to inconsistent future predictions.
This qualitative comparison suggests that the future-state representations learned by CAPE are not only compact, but also encode action-conditioned visual transitions that are essential for modeling future outcomes.

\paragraph{Attention analysis.}
To further understand the internal mechanism of CAPE, we visualize the attention distribution from action queries to visual context tokens in the cross-attention layers of the action-query parallel decoder.
As shown in Fig.~\ref{fig:attn_map}, the cross-attention heatmaps reveal a progressive focusing process.
In early layers, the action queries mainly ground themselves on the acting agent, namely the robotic gripper.
As the network becomes deeper, the attention gradually expands to include the manipulated objects and task-relevant environmental context.
This selective spatial attention helps explain why CAPE remains robust to the heavy visual distractors commonly present in the DROID benchmark.

\begin{table}[!ht]
  \centering
  \normalsize
  \caption{\textbf{Ablation Study of CAPE Designs.} We evaluate the impact of architectural choices and training strategies on the DROID dataset. All variants are trained under identical settings to ensure a fair comparison.}
  \label{tab:ablation}
  \begin{tabular}{lccccc}
    \toprule
    \multirow{2}{*}{\textbf{Variant}} & \multicolumn{3}{c}{\textbf{Retrieval Hit@5 (\%) $\uparrow$}} & \multicolumn{2}{c}{\textbf{Action Score $\uparrow$}} \\ 
    \cmidrule(lr){2-4} \cmidrule(lr){5-6}
    & $h=1$ & $h=3$ & $h=5$ & $h=3$ & $h=5$ \\
    \midrule
    w/o Visual Context Encoder & 86.71 & 82.31 & 78.04 & 48.80 & 46.40 \\
    w/o Negative Action Sample & 91.62 & 86.45 & 82.87 & 56.11 & 52.60  \\
    Action-Visual Alignment & 90.82 & 86.02 & 80.19 & 58.40 & 55.20 \\
    \midrule
    \textbf{Full Model} & \textbf{91.64} & \textbf{87.48} & \textbf{84.19} & \textbf{62.08} & \textbf{56.64} \\ 
    \bottomrule
  \end{tabular}
\end{table}
\subsection{Ablation Study}

We conduct an ablation study on the DROID~\citep{khazatsky2024droid} dataset to evaluate the contribution of each design choice in our framework. 
The results are summarized in Tab.~\ref{tab:ablation}.
Notably, the removal of Visual Context Encoder leads to the most substantial performance degradation, particularly in the Action Score, which drops from 62.08\% to 48.8\% at $h=3$. 
This suggests that, although the frozen DINOv2 encoder provides robust visual representations, the subsequent visual self-attention module offers additional refinement that can better adapt these features to action-conditioned transitions modeling and future state prediction.

Furthermore, training strategies significantly impact the model's learned discriminative representation.
Excluding Negative Action Samples—constructed via batch-wise action swapping—results in a consistent decline in Action Score, underscoring its importance in helping the contrastive framework distinguish between subtle action-conditioned state transitions.
Similarly, the Action-Visual Alignment variant, which simplifies the objective by directly aligning the predictive future state with future observation, underperforms the full model.
This gap demonstrates that our proposed goal-convergent contrastive objective more effectively captures the intricate dependencies between current observations, actions, and their corresponding future outcomes.

\section{Conclusions}

The core contribution of this paper is to reformulate visual dynamics prediction for embodied planning as action-conditioned parallel prediction of future representations, rather than dense autoregressive prediction of future states.
We identify that existing latent world models remain limited by two coupled issues: supervision over high-dimensional future visual states emphasizes action-irrelevant scene redundancy, while autoregressive rollout introduces sequential computation overhead and error accumulation. 
Our proposed CAPE addresses these limitations by using action sequences as temporally ordered queries over a shared visual context and training the resulting future tokens with a goal-convergent contrastive objective. 
Experiments on DROID and RoboCasa show that CAPE improves future-state retrieval, offline action matching, and closed-loop planning, while reducing the planning cost associated with evaluating multi-step candidate action sequences. 
These results suggest that organizing predictive representations around action-induced future differences offers a practical direction for more efficient and scalable model-based embodied planning.

\newpage
\bibliographystyle{plainnat}
\bibliography{nips2026_conference}

@inproceedings{peebles2023adaln,
  title     = {Scalable Diffusion Models with Transformers},
  author    = {Peebles, William and Xie, Saining},
  booktitle = {Proceedings of the IEEE/CVF International Conference on Computer Vision},
  pages     = {4172--4182},
  year      = {2023}
}

@article{oquab2025dinov3,
  title   = {{DINOv3}},
  author  = {Sim{\'e}oni, Oriane and Vo, Huy V. and Seitzer, Maximilian and Baldassarre, Federico and Oquab, Maxime and Jose, Cijo and Khalidov, Vasil and Szafraniec, Marc and Yi, Seungeun and Ramamonjisoa, Micha{\"e}l and Massa, Francisco and Haziza, Daniel and Wehrstedt, Luca and Wang, Jianyuan and Darcet, Timoth{\'e}e and Moutakanni, Th{\'e}o and Sentana, Leonel and Roberts, Claire and Vedaldi, Andrea and Tolan, Jamie and Brandt, John and Couprie, Camille and Mairal, Julien and J{\'e}gou, Herv{\'e} and Labatut, Patrick and Bojanowski, Piotr},
  journal = {arXiv preprint arXiv:2508.10104},
  year    = {2025}
}

@article{survey2025worldmodels,
  title   = {A Comprehensive Survey on World Models for Embodied {AI}},
  author  = {Li, Xinqing and He, Xin and Zhang, Le and Liu, Yun},
  journal = {arXiv preprint arXiv:2510.16732},
  year    = {2025}
}

@article{ai2025review,
  title     = {A Review of Learning-Based Dynamics Models for Robotic Manipulation},
  author    = {Ai, Bo and Tian, Stephen and Shi, Haochen and Wang, Yixuan and Pfaff, Tobias and Tan, Cheston and Christensen, Henrik I. and Su, Hao and Wu, Jiajun and Li, Yunzhu},
  journal   = {Science Robotics},
  volume    = {10},
  number    = {106},
  pages     = {eadt1497},
  year      = {2025}
}

@inproceedings{grill2020byol,
  title     = {Bootstrap Your Own Latent: A New Approach to Self-Supervised Learning},
  author    = {Grill, Jean-Bastien and Strub, Florian and Altch{\'e}, Florent and Tallec, Corentin and Richemond, Pierre H. and Buchatskaya, Elena and Doersch, Carl and Pires, Bernardo Avila and Guo, Zhaohan Daniel and Azar, Mohammad Gheshlaghi and Piot, Bilal and Kavukcuoglu, Koray and Munos, R{\'e}mi and Valko, Michal},
  booktitle = {Advances in Neural Information Processing Systems},
  volume    = {33},
  pages     = {21271--21284},
  year      = {2020}
}

@inproceedings{bardes2022vicreg,
  title     = {{VICReg}: Variance-Invariance-Covariance Regularization for Self-Supervised Learning},
  author    = {Bardes, Adrien and Ponce, Jean and LeCun, Yann},
  booktitle = {International Conference on Learning Representations},
  year      = {2022}
}

@inproceedings{assran2023ijepa,
  title     = {Self-Supervised Learning from Images with a Joint-Embedding Predictive Architecture},
  author    = {Assran, Mahmoud and Duval, Quentin and Misra, Ishan and Bojanowski, Piotr and Vincent, Pascal and Rabbat, Michael and LeCun, Yann and Ballas, Nicolas},
  booktitle = {Proceedings of the IEEE/CVF Conference on Computer Vision and Pattern Recognition},
  pages     = {15619--15629},
  year      = {2023}
}

@inproceedings{laskin2020curl,
  title     = {{CURL}: Contrastive Unsupervised Representations for Reinforcement Learning},
  author    = {Laskin, Michael and Srinivas, Aravind and Abbeel, Pieter},
  booktitle = {Proceedings of the 37th International Conference on Machine Learning},
  pages     = {5639--5650},
  volume    = {119},
  series    = {Proceedings of Machine Learning Research},
  publisher = {PMLR},
  year      = {2020}
}

@inproceedings{schwarzer2021spr,
  title     = {Data-Efficient Reinforcement Learning with Self-Predictive Representations},
  author    = {Schwarzer, Max and Anand, Ankesh and Goel, Rishab and Hjelm, R. Devon and Courville, Aaron and Bachman, Philip},
  booktitle = {International Conference on Learning Representations},
  year      = {2021}
}

@article{oquab2023dinov2,
  title   = {{DINOv2}: Learning Robust Visual Features without Supervision},
  author  = {Oquab, Maxime and Darcet, Timoth{\'e}e and Moutakanni, Th{\'e}o and Vo, Huy V. and Szafraniec, Marc and Khalidov, Vasil and Fernandez, Pierre and Haziza, Daniel and Massa, Francisco and El-Nouby, Alaaeldin and Assran, Mido and Ballas, Nicolas and Galuba, Wojciech and Howes, Russell and Huang, Po-Yao and Li, Shang-Wen and Misra, Ishan and Rabbat, Michael G. and Sharma, Vasu and Synnaeve, Gabriel and Xu, Hu and J{\'e}gou, Herv{\'e} and Mairal, Julien and Labatut, Patrick and Joulin, Armand and Bojanowski, Piotr},
  journal = {Transactions on Machine Learning Research},
  year    = {2024}
}

@inproceedings{hafner2019planet,
  title     = {Learning Latent Dynamics for Planning from Pixels},
  author    = {Hafner, Danijar and Lillicrap, Timothy and Fischer, Ian and Villegas, Ruben and Ha, David and Lee, Honglak and Davidson, James},
  booktitle = {Proceedings of the 36th International Conference on Machine Learning},
  pages     = {2555--2565},
  volume    = {97},
  series    = {Proceedings of Machine Learning Research},
  publisher = {PMLR},
  year      = {2019}
}

@inproceedings{hafner2020dreamer,
  title     = {Dream to Control: Learning Behaviors by Latent Imagination},
  author    = {Hafner, Danijar and Lillicrap, Timothy P. and Ba, Jimmy and Norouzi, Mohammad},
  booktitle = {International Conference on Learning Representations},
  year      = {2020}
}

@inproceedings{bardes2024vjepa,
  title     = {{V-JEPA}: Latent Video Prediction for Visual Representation Learning},
  author    = {Bardes, Adrien and Garrido, Quentin and Ponce, Jean and Chen, Xinlei and Rabbat, Michael and LeCun, Yann and Assran, Mido and Ballas, Nicolas},
  booktitle = {International Conference on Learning Representations},
  year      = {2024}
}

@inproceedings{anand2019unsupervised,
  title     = {Unsupervised State Representation Learning in {Atari}},
  author    = {Anand, Ankesh and Racah, Evan and Ozair, Sherjil and Bengio, Yoshua and C{\^o}t{\'e}, Marc-Alexandre and Hjelm, R. Devon},
  booktitle = {Advances in Neural Information Processing Systems},
  volume    = {32},
  pages     = {8766--8779},
  year      = {2019}
}

@inproceedings{he2020moco,
  title     = {Momentum Contrast for Unsupervised Visual Representation Learning},
  author    = {He, Kaiming and Fan, Haoqi and Wu, Yuxin and Xie, Saining and Girshick, Ross},
  booktitle = {Proceedings of the IEEE/CVF Conference on Computer Vision and Pattern Recognition},
  pages     = {9729--9738},
  year      = {2020}
}

@inproceedings{chen2020simclr,
  title     = {A Simple Framework for Contrastive Learning of Visual Representations},
  author    = {Chen, Ting and Kornblith, Simon and Norouzi, Mohammad and Hinton, Geoffrey},
  booktitle = {Proceedings of the 37th International Conference on Machine Learning},
  pages     = {1597--1607},
  volume    = {119},
  series    = {Proceedings of Machine Learning Research},
  publisher = {PMLR},
  year      = {2020}
}

@article{oord2018cpc,
  title   = {Representation Learning with Contrastive Predictive Coding},
  author  = {van den Oord, A{\"a}ron and Li, Yazhe and Vinyals, Oriol},
  journal = {arXiv preprint arXiv:1807.03748},
  year    = {2018}
}

@inproceedings{finn2016unsupervised,
  title     = {Unsupervised Learning for Physical Interaction through Video Prediction},
  author    = {Finn, Chelsea and Goodfellow, Ian and Levine, Sergey},
  booktitle = {Advances in Neural Information Processing Systems},
  volume    = {29},
  pages     = {64--72},
  year      = {2016}
}

@inproceedings{oh2015action,
  title     = {Action-Conditional Video Prediction Using Deep Networks in {Atari} Games},
  author    = {Oh, Junhyuk and Guo, Xiaoxiao and Lee, Honglak and Lewis, Richard L. and Singh, Satinder},
  booktitle = {Advances in Neural Information Processing Systems},
  volume    = {28},
  pages     = {2863--2871},
  year      = {2015}
}

@inproceedings{finn2017deep,
  title        = {Deep Visual Foresight for Planning Robot Motion},
  author       = {Finn, Chelsea and Levine, Sergey},
  booktitle    = {IEEE International Conference on Robotics and Automation},
  pages        = {2786--2793},
  year         = {2017}
}

@article{ebert2018visual,
  title   = {Visual Foresight: Model-Based Deep Reinforcement Learning for Vision-Based Robotic Control},
  author  = {Ebert, Frederik and Finn, Chelsea and Dasari, Sudeep and Xie, Annie and Lee, Alex and Levine, Sergey},
  journal = {arXiv preprint arXiv:1812.00568},
  year    = {2018}
}

@article{dasari2019robonet,
  title   = {{RoboNet}: Large-Scale Multi-Robot Learning},
  author  = {Dasari, Sudeep and Ebert, Frederik and Tian, Stephen and Nair, Suraj and Bucher, Bernadette and Schmeckpeper, Karl and Singh, Siddharth and Levine, Sergey and Finn, Chelsea},
  journal = {arXiv preprint arXiv:1910.11215},
  year    = {2019}
}

@inproceedings{ebert2017self,
  title     = {Self-Supervised Visual Planning with Temporal Skip Connections},
  author    = {Ebert, Frederik and Finn, Chelsea and Lee, Alex X. and Levine, Sergey},
  booktitle = {Proceedings of the 1st Annual Conference on Robot Learning},
  pages     = {344--356},
  volume    = {78},
  series    = {Proceedings of Machine Learning Research},
  publisher = {PMLR},
  year      = {2017}
}

@inproceedings{fragkiadaki2016learning,
  title     = {Learning Visual Predictive Models of Physics for Playing Billiards},
  author    = {Fragkiadaki, Katerina and Agrawal, Pulkit and Levine, Sergey and Malik, Jitendra},
  booktitle = {International Conference on Learning Representations},
  year      = {2016}
}

@inproceedings{williams2017information,
  title        = {Information Theoretic {MPC} for Model-Based Reinforcement Learning},
  author       = {Williams, Grady and Wagener, Nolan and Goldfain, Brian and Drews, Paul and Rehg, James M. and Boots, Byron and Theodorou, Evangelos A.},
  booktitle    = {IEEE International Conference on Robotics and Automation},
  pages        = {1714--1721},
  year         = {2017}
}

@inproceedings{khazatsky2024droid,
  title     = {{DROID}: A Large-Scale In-the-Wild Robot Manipulation Dataset},
  author    = {Khazatsky, Alexander and Pertsch, Karl and Nair, Suraj and Balakrishna, Ashwin and Dasari, Sudeep and Karamcheti, Siddharth and Nasiriany, Soroush and others},
  booktitle = {Robotics: Science and Systems},
  year      = {2024}
}

@inproceedings{nasiriany2024robocasa,
  title     = {{RoboCasa}: Large-Scale Simulation of Everyday Tasks for Generalist Robots},
  author    = {Nasiriany, Soroush and Maddukuri, Abhiram and Zhang, Lance and Parikh, Adeet and Lo, Aaron and Joshi, Abhishek and Mandlekar, Ajay and Zhu, Yuke},
  booktitle = {Robotics: Science and Systems},
  year      = {2024}
}

@article{zhou2024dinowm,
  title   = {{DINO-WM}: World Models on Pre-trained Visual Features enable Zero-shot Planning},
  author  = {Zhou, Gaoyue and Pan, Hengkai and LeCun, Yann and Pinto, Lerrel},
  journal = {arXiv preprint arXiv:2411.04983},
  year    = {2024}
}

@inproceedings{zheng2023taco,
  title     = {{TACO}: Temporal Latent Action-Driven Contrastive Loss for Visual Reinforcement Learning},
  author    = {Zheng, Ruijie and Wang, Xiyao and Sun, Yanchao and Ma, Shuang and Zhao, Jieyu and Xu, Huazhe and Daum{\'e} III, Hal and Huang, Furong},
  booktitle = {Advances in Neural Information Processing Systems},
  volume    = {36},
  pages     = {48203--48225},
  year      = {2023}
}

@inproceedings{zheng2024premiertaco,
  title     = {{Premier-TACO} is a Few-Shot Policy Learner: Pretraining Multitask Representation via Temporal Action-Driven Contrastive Loss},
  author    = {Zheng, Ruijie and Liang, Yongyuan and Wang, Xiyao and Ma, Shuang and Daum{\'e} III, Hal and Xu, Huazhe and Langford, John and Palanisamy, Praveen and Basu, Kalyan Shankar and Huang, Furong},
  booktitle = {Proceedings of the 41st International Conference on Machine Learning},
  pages     = {61413--61431},
  volume    = {235},
  series    = {Proceedings of Machine Learning Research},
  publisher = {PMLR},
  year      = {2024}
}

@article{hafner2023dreamerv3,
  title   = {Mastering Diverse Control Tasks through World Models},
  author  = {Hafner, Danijar and Pasukonis, Jurgis and Ba, Jimmy and Lillicrap, Timothy},
  journal = {Nature},
  volume  = {640},
  pages   = {647--653},
  year    = {2025}
}

@inproceedings{micheli2023IRIS,
  title     = {Transformers Are Sample-Efficient World Models},
  author    = {Micheli, Vincent and Alonso, Eloi and Fleuret, Fran{\c{c}}ois},
  booktitle = {International Conference on Learning Representations},
  year      = {2023}
}

@article{assran2025vjepa2,
  title   = {{V-JEPA} 2: Self-Supervised Video Models Enable Understanding, Prediction and Planning},
  author  = {Assran, Mido and others},
  journal = {arXiv preprint arXiv:2506.09985},
  year    = {2025}
}

@article{terver2025jepa-wm,
  title   = {What Drives Success in Physical Planning with Joint-Embedding Predictive World Models?},
  author  = {Terver, Basile and Yang, Tsung-Yen and Ponce, Jean and Bardes, Adrien and LeCun, Yann},
  journal = {arXiv preprint arXiv:2512.24497},
  year    = {2025}
}

@inproceedings{ha2018world,
  title     = {Recurrent World Models Facilitate Policy Evolution},
  author    = {Ha, David and Schmidhuber, J{\"u}rgen},
  booktitle = {Advances in Neural Information Processing Systems},
  volume    = {31},
  pages     = {2455--2467},
  year      = {2018}
}

@inproceedings{kaiser2020model,
  title     = {Model-Based Reinforcement Learning for {Atari}},
  author    = {Kaiser, Lukasz and Babaeizadeh, Mohammad and Milos, Piotr and Osi{\'n}ski, B{\l}a{\.z}ej and Campbell, Roy H. and Czechowski, Krzysztof and Erhan, Dumitru and Finn, Chelsea and Kozakowski, Piotr and Levine, Sergey and others},
  booktitle = {International Conference on Learning Representations},
  year      = {2020}
}

@inproceedings{kipf2020contrastive,
  title     = {Contrastive Learning of Structured World Models},
  author    = {Kipf, Thomas and van der Pol, Elise and Welling, Max},
  booktitle = {International Conference on Learning Representations},
  year      = {2020}
}

@inproceedings{hafner2021dreamerv2,
  title     = {Mastering {Atari} with Discrete World Models},
  author    = {Hafner, Danijar and Lillicrap, Timothy P. and Norouzi, Mohammad and Ba, Jimmy},
  booktitle = {International Conference on Learning Representations},
  year      = {2021}
}

@inproceedings{seo2023masked,
  title     = {Masked World Models for Visual Control},
  author    = {Seo, Younggyo and Hafner, Danijar and Liu, Hao and Liu, Fangchen and James, Stephen and Lee, Kimin and Abbeel, Pieter},
  booktitle = {Proceedings of the 6th Conference on Robot Learning},
  pages     = {1332--1344},
  volume    = {205},
  series    = {Proceedings of Machine Learning Research},
  publisher = {PMLR},
  year      = {2023}
}

@article{guo2026flowdreamer,
  title   = {{FlowDreamer}: An {RGB-D} World Model with Flow-Based Motion Representations for Robot Manipulation},
  author  = {Guo, Jun and Ma, Xiaojian and Wang, Yikai and Yang, Min and Liu, Huaping and Li, Qing},
  journal = {arXiv preprint arXiv:2505.10075},
  year    = {2025}
}

@article{liu2025embodied,
  title   = {Aligning Cyber Space with Physical World: A Comprehensive Survey on Embodied {AI}},
  author  = {Liu, Yang and Chen, Weixing and Bai, Yongjie and Liang, Xiaodan and Li, Guanbin and Gao, Wen and Lin, Liang},
  journal = {IEEE/ASME Transactions on Mechatronics},
  volume  = {30},
  number  = {6},
  pages   = {7253--7274},
  year    = {2025}
}

@article{garcia1989mpc,
  title     = {Model Predictive Control: Theory and Practice---A Survey},
  author    = {Garcia, Carlos E. and Prett, David M. and Morari, Manfred},
  journal   = {Automatica},
  volume    = {25},
  number    = {3},
  pages     = {335--348},
  year      = {1989}
}

@article{ferraro2025focus,
  title   = {{FOCUS}: Object-Centric World Models for Robotic Manipulation},
  author  = {Ferraro, Stefano and Mazzaglia, Pietro and Verbelen, Tim and Dhoedt, Bart},
  journal = {Frontiers in Neurorobotics},
  volume  = {19},
  pages   = {1585386},
  year    = {2025}
}

@article{zhang2025objects,
  title   = {Objects Matter: Object-Centric World Models Improve Reinforcement Learning in Visually Complex Environments},
  author  = {Zhang, Weipu and Jelley, Adam and McInroe, Trevor and Storkey, Amos},
  journal = {arXiv preprint arXiv:2501.16443},
  year    = {2025}
}

@inproceedings{zhang2023storm,
  title     = {{STORM}: Efficient Stochastic Transformer-Based World Models for Reinforcement Learning},
  author    = {Zhang, Weipu and Wang, Gang and Sun, Jian and Yuan, Yetian and Huang, Gao},
  booktitle = {Advances in Neural Information Processing Systems},
  volume    = {36},
  pages     = {27147--27166},
  year      = {2023}
}

@inproceedings{bruce2024genie,
  title     = {{Genie}: Generative Interactive Environments},
  author    = {Bruce, Jake and Dennis, Michael and Edwards, Ashley and Parker-Holder, Jack and Shi, Yuge and Hughes, Edward and Lai, Matthew and Mavalankar, Aditi and Steigerwald, Richie and Apps, Chris and others},
  booktitle = {Proceedings of the 41st International Conference on Machine Learning},
  volume    = {235},
  pages     = {4603--4623},
  series    = {Proceedings of Machine Learning Research},
  publisher = {PMLR},
  year      = {2024}
}


\newpage

\appendix
{\LARGE\bf Appendix}

The supplementary material is organized as follows. 
Sec.~\ref{sec:app-a} describes the implementation details of the CAPE architecture and training hyperparameters; 
Sec.~\ref{sec:app-b} specifies the baseline configurations, including task-specific adaptations and alignment across baselines.
Sec.~\ref{sec:app-c} elaborates on the evaluation protocols for future state retrieval, DROID~\citep{khazatsky2024droid} offline action matching, and RoboCasa~\citep{nasiriany2024robocasa} zero-shot planning.
Sec.~\ref{sec:app-d} provides additional qualitative visualizations,
and Sec.~\ref{sec:app-e} discusses limitations and future directions.
Code, pretrained models, and evaluation environments will be released upon publication.

\section{Implementation Details}
\label{sec:app-a}

\paragraph{Training procedure.} 
All models are implemented in PyTorch.
The observation encoder is initialized from a pre-trained DINOv2~\citep{oquab2023dinov2} ViT-S/14 model loaded from the HuggingFace Hub, and its weights are kept frozen throughout all experiments. 
The visual context encoder takes the output tokens from the frozen DINOv2~\citep{oquab2023dinov2} encoder as input and consists of three self-attention layers. 
The action encoder is a linear layer that projects each raw action vector into a $d$-dimensional query token, where $d=384$ matches the DINOv2~\citep{oquab2023dinov2} feature dimension. 
The parallel action-query decoder consists of six stacked layers, each comprising causal self-attention over action queries and cross-attention between action queries and visual context tokens. 

CAPE is trained on 4$\times$ NVIDIA L40 GPUs using the AdamW optimizer with a cosine annealing learning rate schedule for 80 epochs on the 8,000-trajectory DROID~\citep{khazatsky2024droid} subset.
Detailed training and architectural hyperparameters are provided in Table~\ref{tab:training-hyperparameters} and Table~\ref{tab:arch_hyperparams}, respectively. 
Each training batch contains $n_{\text{group}}=16$ groups, where all samples within a group are drawn from the same episode, ensuring that hard negatives share identical visual background and task context while differing only in their action trajectories. 
The contrastive temperature $\tau=0.07$ is fixed throughout training and not tuned.

\begin{table}[!htbp]
\centering
\caption{Training hyperparameters.}
\label{tab:training-hyperparameters}
\begin{tabular}{ll}
\toprule
Hyperparameter & Value \\
\midrule
Optimizer & AdamW \\
Learning Rate & $1 \times 10^{-4}$ \\
Weight Decay & $1 \times 10^{-4}$ \\
LR scheduler & CosineAnnealingLR \\
Batch size & 2048 \\
Number of epochs & 80 \\
Samples per group & 16 \\
Contrastive temperature ($\tau$) & 0.07 \\
Horizon (prediction length) & 5 \\
\bottomrule
\end{tabular}
\end{table}

\begin{table}[!htbp]
\centering
\caption{Architectural Hyperparameters}
\label{tab:arch_hyperparams}
\begin{tabular}{ll}
\toprule
\textbf{Hyperparameter} & \textbf{Value} \\ 
\midrule
Vision Encoder & DINOv2-ViT-S/14 \\
Action-query Dim & 384 \\
Self-Attention Layers & 3 \\
Cross-Attention Layers & 6 \\
Attention Heads & 8 \\
MLP Ratio & 4 \\ 
\bottomrule
\end{tabular}
\end{table}

\paragraph{Empty-action state encoding.}
CAPE uses the same action-query decoding pathway for both action-conditioned prediction and action-free state encoding. 
Given a current observation and an action sequence, each action is projected into an action query, which interacts with the visual context tokens through causal self-attention and cross-attention to produce the predicted future state token. 
When encoding a state without executing any action, we replace the action query with a learnable empty-action query while keeping the rest of the forward computation unchanged. 
This empty-action query represents the state token of an observation under a zero-action transition. 
We use this empty-action state encoding for candidate-key encoding in retrieval, goal-state encoding in planning, and zero-horizon training samples where the action sequence length is zero. 
This design ensures that predicted future state tokens and action-free state tokens are embedded in the same representation space.

\begin{algorithm}[H]
\caption{Cross-Entropy Method}
\label{alg:cem}
\begin{algorithmic}[1]
\STATE $\mu^0 \in \mathbb{R}^{H \times A}$ is zero and covariance matrix $\sigma^0 \text{I} \in \mathbb{R}^{(H \times A)^2}$ is the identity. Number of optimisation steps $J$.
\FOR{$j = 1$ to $J$}
    \STATE Sample $N$ independent trajectories $(\{a_t, \dots, a_{t+H-1}\}) \sim \mathcal{N}(\mu^j, (\sigma^j)^2 \text{I})$
    \STATE For each of the $N$ trajectories, unroll predictor to predict the resulting trajectory, $\hat{z}_i = P_\theta(\hat{z}_{i-1}, a_{i-1}), i = t+1, \dots, t+H$. Compute cost $L^p_\alpha(o_t, a_{t:t+H-1}, o_g)$ for each candidate trajectory.
    \STATE Select top $K$ action sequences with the lowest cost, denote them $(\{a_t, \dots, a_{t+H-1}\})_{1,\dots,K}$. Update
    \begin{equation*}
        \mu^{j+1} = \frac{1}{K} \sum_{k=1}^{K} (\{a_t, \dots, a_{t+H-1}\})_k
    \end{equation*}
    \begin{equation*}
        \sigma^{j+1} = \sqrt{\frac{1}{K-1} \sum_{k=1}^{K} [(\{a_t, \dots, a_{t+H-1}\})_k - \mu^{j+1}]^2}
    \end{equation*}
\ENDFOR
\STATE Step the first $m$ actions of $\mu^J$. where $m \le H$ is a planning hyperparameter in the environment. If we are in MPC mode, the process then repeats at the next time step with the new context observation.
\end{algorithmic}
\end{algorithm}

\paragraph{Planning procedure.} 
At test time, CAPE is deployed within a CEM-based MPC loop. 
Algorithm~\ref{alg:cem} provides the detailed CEM planning procedure.
Given an initial observation and a goal image, the planner samples $N$ candidate action sequences at each iteration and performs $J$ optimization iterations. 
For each candidate trajectory, CAPE predicts the final state token $\hat{u}_{t+H}$ conditioned on the initial observation and the sampled action sequence. 
The planning cost is computed as the cosine distance between $\hat{u}_{t+H}$ and the goal state token $u_g$. 
The goal state token $u_g$ is obtained from the goal image using the empty-action state encoding described above.

\paragraph{Visual Decoder.} 
To qualitatively interpret the learned future predictive representations, we fine-tune a visual decoder to reconstruct future observations from predicted latent tokens. 
Specifically, we initialize the decoder with the weights of the ViT-S decoder with depth 12 released by JEPA-WM~\citep{terver2025jepa-wm}. 
Different from the original design, which reconstructs images directly from DINO features, our decoder takes the DINO feature of the current observation as input and is additionally conditioned on the predicted future state token produced by CAPE's action-query mechanism.
The predicted future state token is injected into the decoder through adaptive layer normalization (AdaLN~\citep{peebles2023adaln}), allowing the reconstruction process to be modulated by the action-induced latent transition.
The decoder is fine-tuned to synthesize the corresponding future frame and is trained with a combination of pixel-space $\ell_2$ loss and perceptual loss. 
We emphasize that the decoder is used solely for visualization and does not affect either the training or inference of CAPE. 
This setup allows us to analyze the information captured by the action-query tokens in the form of reconstructed future observations.

\section{Baseline Configurations}
\label{sec:app-b}

\begin{table}[ht]
\centering
\caption{Model Comparison: Parameter Counts}
\label{tab:parameter}
\resizebox{\textwidth}{!}{%
\begin{tabular}{@{}lllll@{}}
\toprule
\textbf{Model Category} & \textbf{Method} & \textbf{Visual Encoder (M)} & \textbf{Dynamics Model (M)} & \textbf{Total (M)} \\ \midrule
\multirow{2}{*}{Generative / Discrete} 
 & IRIS & 12.5M & 19.2M & 31.7M  \\
 & DreamerV3 & 17M & 20.74M  & 37.74M  \\ \midrule
Temporal Contrastive 
 & Premier-TACO & 23.5M & 2.38M & 25.88M  \\ \midrule
\multirow{3}{*}{Latent Predictive} 
 & DINO-WM & 21M (Frozen) & 19.3M & 40.3M\\
 & JEPA-WM & 300M (Frozen) & 228.84M & 528.84M \\ 
 & V-JEPA-2-AC & 1000M(Frozen) & 305.21M & 1305.21M \\ \midrule
Ours 
 & CAPE & 21M (Frozen) & 21.3M & 42.3M \\ \bottomrule
\end{tabular}%
}
\end{table}

To ensure a rigorous and fair evaluation, we align both the training protocols and parameter capacities across all evaluated latent world models. 
Table~\ref{tab:parameter} summarizes the parameter distributions of CAPE and the selected baselines, grouped according to their underlying architectural paradigms. 
We select these baselines because each of them contains an explicit or implicit world-modeling component, making them suitable for evaluating the quality of learned future representations and the computational efficiency of action-conditioned future prediction.

\textbf{Latent Predictive Baselines.} 
As discussed in the main text, CAPE shares a closely related latent predictive paradigm with DINO-WM~\citep{zhou2024dinowm}, JEPA-WM~\citep{terver2025jepa-wm}, and V-JEPA-2-AC~\citep{assran2025vjepa2}, where future dynamics are modeled in a learned visual representation space rather than directly in pixel space. 
For these baselines, we follow the evaluation protocols of JEPA-WM~\citep{terver2025jepa-wm} and use the corresponding official open-source checkpoints pre-trained on DROID~\citep{khazatsky2024droid}. 
Specifically, DINO-WM~\citep{zhou2024dinowm}, JEPA-WM~\citep{terver2025jepa-wm}, and V-JEPA-2-AC~\citep{assran2025vjepa2} employ frozen DINOv2-S (21M), DINOv3-L (300M), and V-JEPA2 (1000M) vision encoders, respectively. 

\textbf{Generative and Temporal Contrastive Baselines (Alignment Strategy).} 
Unlike the latent predictive baselines, IRIS~\citep{micheli2023IRIS}, DreamerV3~\citep{hafner2023dreamerv3}, and Premier-TACO~\citep{zheng2024premiertaco} were originally developed primarily in reinforcement learning settings, where model learning is typically coupled with task-specific information such as rewards, policies, or online interaction. 
To isolate and evaluate their world-modeling capability in our setting, we adapt only their world-model components and train them entirely offline on DROID~\citep{khazatsky2024droid}, without using reward labels, task-specific success signals, or online environment interaction. 
This adaptation enables a direct comparison of their ability to model action-conditioned visual transitions from real-world robot trajectories.

Crucially, to ensure a fair comparison and mitigate the disadvantage of not utilizing massive pre-trained foundation models, we scaled up the parameter capacities of IRIS~\citep{micheli2023IRIS}, DreamerV3~\citep{hafner2023dreamerv3}, and Premier-TACO~\citep{zheng2024premiertaco} beyond their default configurations (as reflected in Table~\ref{tab:parameter}). 
Furthermore, to align with the training distribution established by the latent predictive baselines JEPA-WM~\citep{terver2025jepa-wm}, we trained these models from scratch using an identical subset of the DROID dataset. 
This subset comprises exactly 8,000 trajectories, with sequence lengths strictly ranging from 20 to 50 steps. 
All models trained from scratch use 1--4 NVIDIA L40 GPUs depending on model capacity.
This alignment ensures that any performance discrepancies observed in our experiments stem fundamentally from architectural inductive biases (e.g., autoregressive vs. parallel decoding, pixel reconstruction vs. latent contrastive) rather than disparities in model capacity or training data volume.

\section{Experiments Settings}
\label{sec:app-c}

\subsection{Future State Retrieval}
We evaluate the predictive fidelity of learned representations via a future state retrieval task on the held-out DROID~\citep{khazatsky2024droid} dataset, which was originally recorded at 15 FPS. 
We apply a temporal stride of 6 to the video sequences, resulting in a physical time interval of 0.4 seconds per predictive step.
Given a current observation $o_t$ and a sequence of future actions $\mathbf{a}_{t:t+H-1}$, each model is required to predict the future state representation $\hat{z}_{t+H}$ and retrieve the corresponding  ground-truth future frame from a candidate pool constructed from the same held-out split. 

Specifically, the held-out split contains 800 episodes that are not used for training, with each episode consisting of 20--50 temporally strided steps.
For each query, the model is asked to identify the correct future state from the same episode and the same camera view.
The candidate pool is constructed using all frames from 10 episodes of the same task, yielding approximately 420 competitive candidates under the same environment. 
We use Hit@K as the quantitative retrieval metric, defined as the percentage of queries for which the ground-truth future state is ranked among the top-$K$ candidates according to the predicted-state similarity. 
We evaluate three prediction horizons, $t+1$ $(0.4\mathrm{s})$, $t+3$ $(1.2\mathrm{s})$, and $t+5$ $(2.0\mathrm{s})$, and report Hit@1, Hit@5, and Hit@10 at each horizon to assess both short- and long-term dynamics modeling capability.

To ensure a fair comparison, we made minimal yet reasonable modifications to each baseline to adapt it to the retrieval task. 
For CAPE, retrieval queries are decoded from action query conditioned on the visual context token of the current observation, while candidate keys are obtained from candidate frame observations using the learnable empty-action query.
Retrieval is then performed by computing the cosine similarity between the predicted compact single-token representation and the encoded candidate frames. 
For baselines that produce dense spatial token predictions, including DINO-WM~\citep{zhou2024dinowm}, V-JEPA2-AC~\citep{assran2025vjepa2}, JEPA-WM~\citep{terver2025jepa-wm}, and IRIS~\citep{micheli2023IRIS}, we directly use the same distance metric adopted in their respective MPC planning phases.
Specifically, we compute the L2 distance between the predicted patch tokens and the candidate patch tokens to rank candidate frames, since both retrieval and MPC planning require measuring the cost between a predicted state and a target state. 
For baselines that represent the state as a single latent vector, including DreamerV3~\citep{hafner2023dreamerv3} and Premier-TACO~\citep{zheng2024premiertaco}, we compute the L2 distance between candidate keys and retrieval queries.

\subsection{DROID Offline Evaluation}
\paragraph{Offline Action Matching}
To evaluate models without having to step actions in a real robot or a simulator, we compare the actions output by the planner and the groundtruth actions of the trajectory sampled from the dataset to define initial and goal state. 
Given an initial and goal observation pair sampled from the held-out DROID~\citep{khazatsky2024droid} dataset, we run the MPC planner and compute the L1 error between the actions outputted by the planner and the ground-truth actions of the corresponding trajectory. 
We adopt the Action Score metric as defined in JEPA-WM~\citep{terver2025jepa-wm}.
The Action Score is defined as a rescaled inverse of this error: $ 800 \times (0.1 - \mathcal{E}) $ if $ \mathcal{E} < 0.1$ , else 0, where $ \mathcal{E}$ denotes the mean $L1$ action error on the first three dimensions corresponding to end-effector position control, which is more relevant for the tasks we consider. 
We evaluate on 50 held-out episodes sampled from DROID~\citep{khazatsky2024droid}, none of which are used for training, focusing on object interaction and arm navigation scenarios. 
Following JEPA-WM~\citep{terver2025jepa-wm}, actions are defined as deltas in measured robot positions and are not normalized. 
We plan with CEM at horizon $H=3$, with $N=300$ candidate trajectories, $J=15$ iterations, and a planning context window $W_p=1$.

\paragraph{MPC Planning Efficiency.}

\begin{table}[!htbp]
\centering
\caption{\textbf{Environment-specific hyperparameters for planning}. 
$N$ is the number of trajectories evaluated in parallel, $H$ the planning horizon, $m$ the number of actions to step in the environment, $K$ the number of top actions in CEM, $J$ the number of iterations of the optimizer, $M$ the number of steps per planning episode. }
\label{tab:cem-hyperparameters}
\begin{tabular}{lcccccc}
\toprule
Hyperparameter & $N$ & $H$ & $m$ & $K$ & $J$ & $M$ \\
\midrule
DROID & 300 & 1/3/5 & 1/3/5 & 10 & 15 & - \\
RoboCasa & 300 & 3/5 & 1 &  10 & 15 & 60\\
\bottomrule
\end{tabular}
\end{table}
We evaluate the planning efficiency of each latent world model by measuring the runtime of a single MPC decision loop. 
For all models, we adopt a CEM-based MPC planner with an identical hyperparameter configuration, as reported in Tab.~\ref{tab:cem-hyperparameters}. 
To isolate the computational overhead of latent-space planning, image encoding time is excluded from the measurement. 
The measured runtime therefore covers only the forward prediction of future latent states for sampled action trajectories and the computation of MPC costs between the predicted future states and the goal state.
All runtimes are measured on a single NVIDIA RTX 4090 GPU with batch size 2 and averaged over all evaluation queries.

\subsection{RoboCasa Zero-Shot Closed-Loop Planning}
We evaluate zero-shot generalization by directly applying our DROID-trained model to the RoboCasa~\citep{nasiriany2024robocasa} simulation environment without any fine-tuning. 
We benchmark on two custom pick-and-place subtasks extracted from teleoperated trajectories: $\textbf{Reach}$ (Rc-R) , where success requires the end-effector to reach within 0.2 simulator units ($\approx 5 cm$) of the target object, and $\textbf{Place}$ (Rc-Pl), where success requires the object to be placed within 0.15 simulator units of the target position. 
The goal frame is taken as the last frame of a teleoperated trajectory. 
Each evaluation set consists of 16 teleoperated trajectories in a kitchen scene with varied object categories. 
For each subtask, we evaluate 16 goal-conditioned trials per evaluation seed and report results averaged over $5$ seeds, using the final frame of each teleoperated trajectory as the goal frame. 
To account for the control frequency mismatch between DROID~\citep{khazatsky2024droid} (2.5 fps) and RoboCasa~\citep{nasiriany2024robocasa} (20 Hz), we apply 8 $\times$ action repeat.
Planner actions are rescaled to match the RoboCasa~\citep{nasiriany2024robocasa} action space: $ [-1,1] $ for end-effector delta position and $[-1,1] $ for gripper closure. 
We use the same Franka Panda gripper as in DROID~\citep{khazatsky2024droid}. 
We plan with CEM at horizon $H=5$, with $N=300$ candidate trajectories, $J=15$ iterations, and $m=1$ action stepped per replanning cycle, for a total of 60 replanning steps per episode.

\section{Additional Results}
\label{sec:app-d}

\subsection{Qualitative Visualization Analysis}
\begin{figure*}[t]
  \centering
  \includegraphics[width=\textwidth]{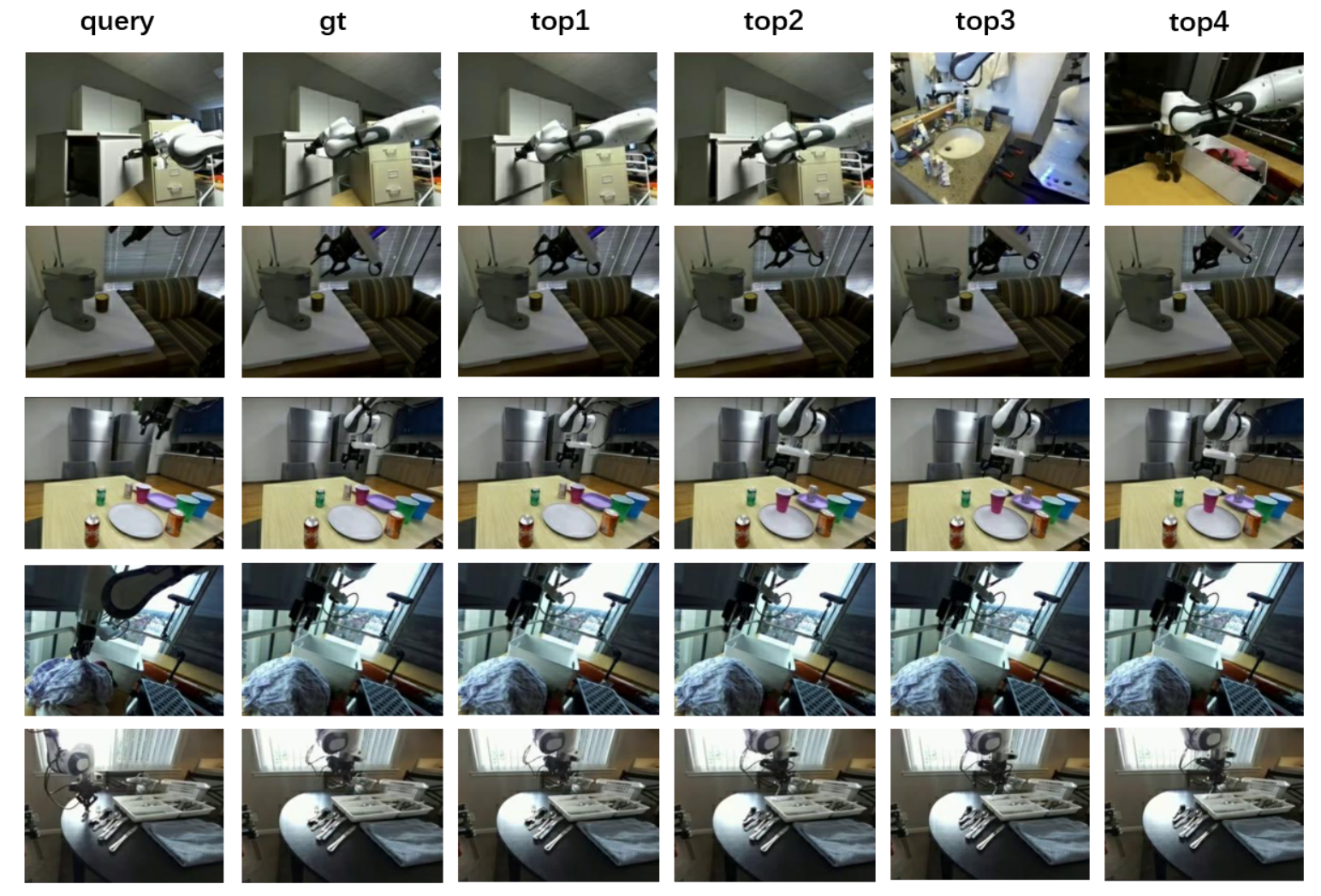}
  \caption{
    \textbf{Future state retrieval visualization for horizon } $h=5$ (2.0s) .
    Given the input query frame and action sequence (left), CAPE (Ours) accurately predicts the future action-driven state transition and retrieves the Ground Truth frame and relevant candidates. Baseline models fail to capture the long-term dynamics, retrieving frames close to the starting configuration.
  }
  \label{fig:retrieval_vis}
\end{figure*}

\paragraph{Results of Long-Horizon Retrieval}
We visualize the top-4 retrieved future frames for $h=5$ (2.0s) of our model in Figure~\ref{fig:retrieval_vis}. 
It shows that our model can infer the right future state based on current observation and action.

\begin{figure}[h!]
  \centering
  \includegraphics[width=1.00\linewidth]{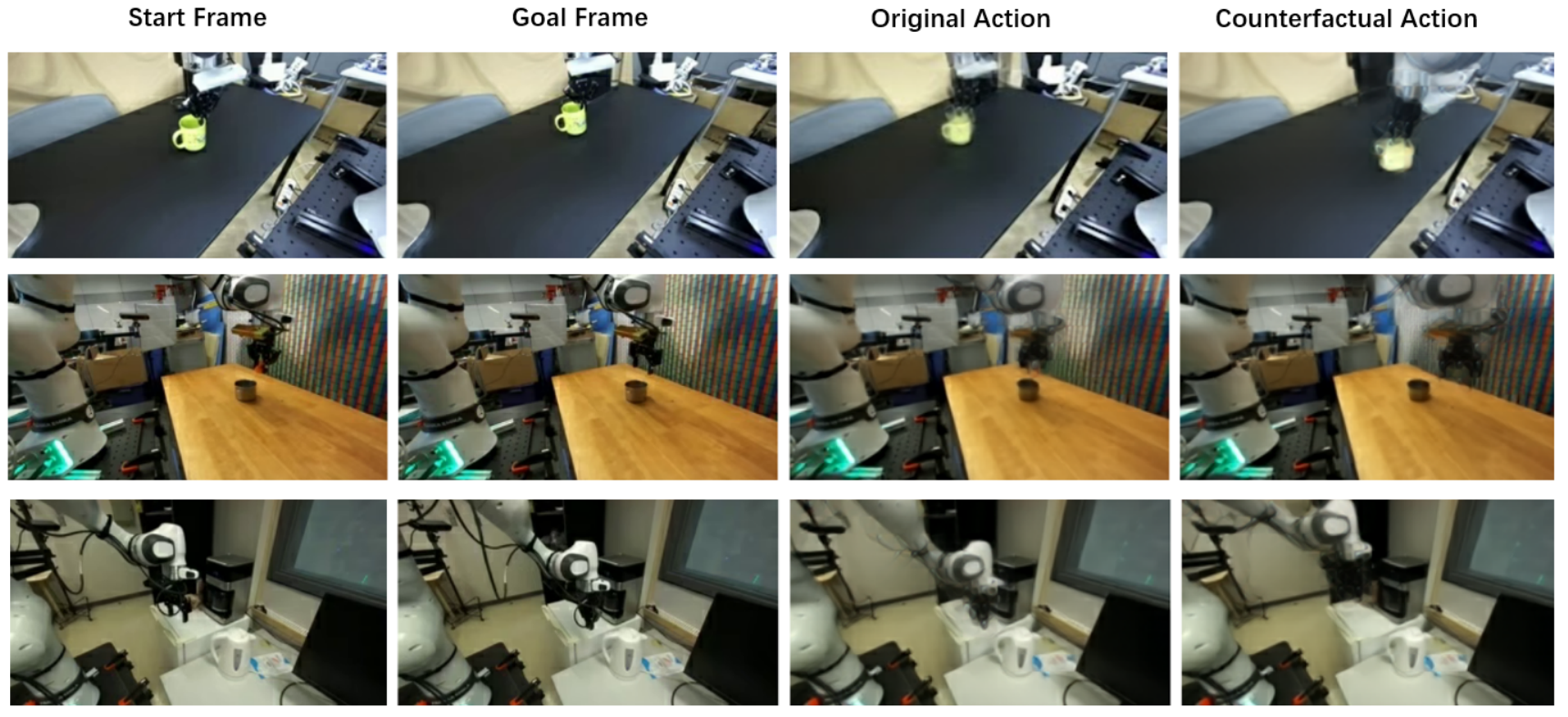}
  \caption{\textbf{Visualization of action-conditioned predicted future tokens.}
  Visualization of action queries based on initial observation across diverse real-world scenes.
  The ``Original Action'' column visualizes future states reconstructed from latent tokens predicted using the actual action sequences executed in the dataset.
  The ``Counterfactual Action'' column presents predictions conditioned on hardcoded Cartesian translations along specific axes ($-x$, $-y$, and $z$). 
  }
  \label{fig:recons}
\end{figure}

\begin{figure}[h!]
  \centering
  \includegraphics[width=1.00\linewidth]{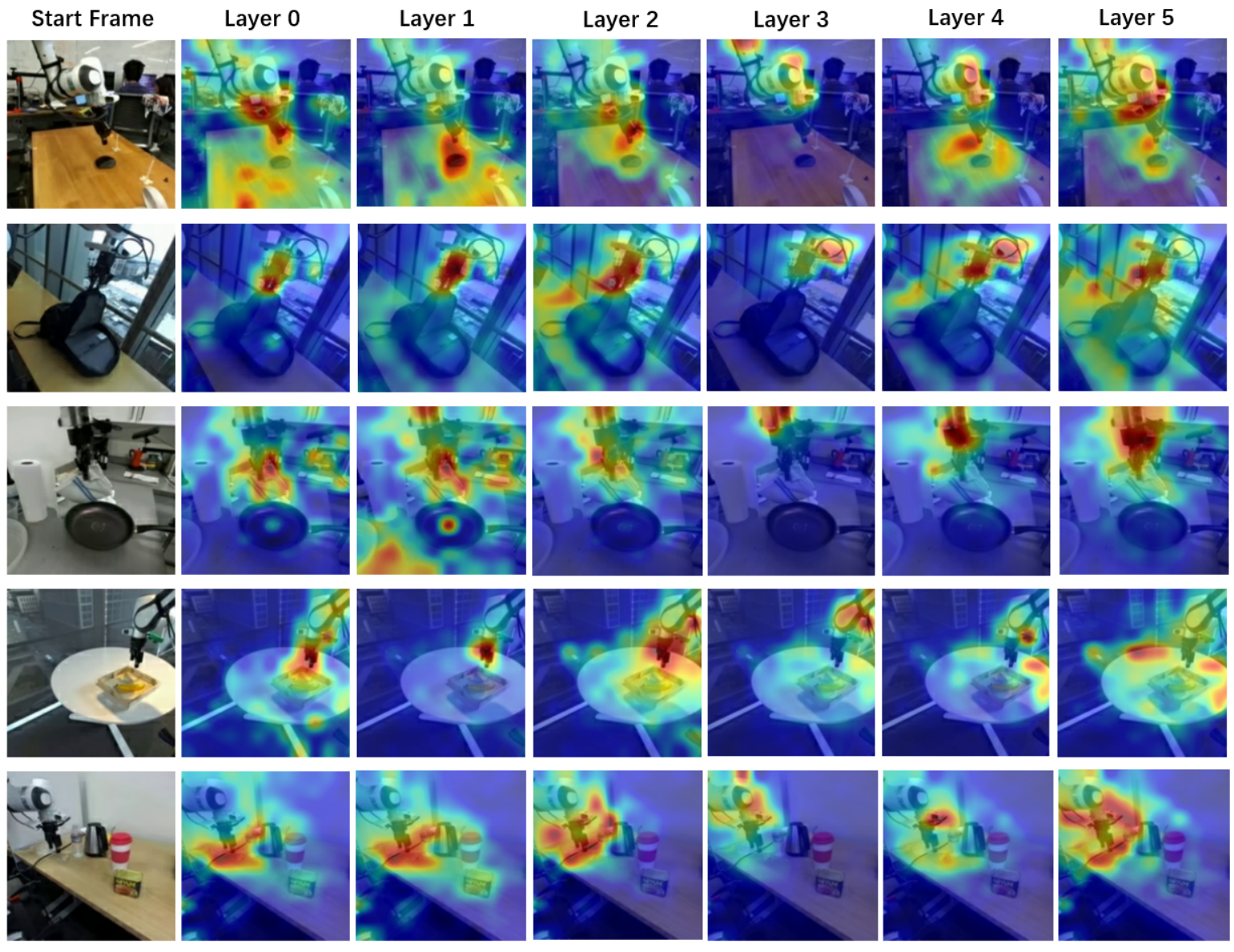}
  \caption{\textbf{Full-depth visualization of action-query attention over visual context.}
  We visualize the cross-attention heatmaps from the action query to the spatial visual context tokens of the input start frame $o_t$ across all six cross-attention layers.
  }
  \label{fig:full_attn}
\end{figure}

\paragraph{Visualization of Model with Counterfactual Action}
Fig.~\ref{fig:recons} presents reconstructions conditioned on both actual executed actions and hypothetical hardcoded counterfactual actions.
Comparing the ``Original Action'' predictions with the groundtruth demonstrates that CAPE accurately captures agent's action in visual observation.
Furthermore, to verify the model’s capability for counterfactual prediction, we observe that the model’s behaviors under hardcoded action inputs (Col4) are consistent with the correct action motions.
This causal alignment confirms that the state tokens are strictly driven by action signals rather than 
observational shortcuts.

\paragraph{Visualization of action-query attention over visual context}
Fig.~\ref{fig:full_attn} presents full-depth visualizations of the cross-attention maps from the action query to the spatial visual context tokens across all six cross-attention layers in the parallel action-query decoder. 
Across different scenes and decoder layers, the action query selectively attends to regions that are most relevant for predicting action-conditioned visual transitions, such as the robot arm, the gripper, and potentially interactable objects. 
In contrast, static or task-irrelevant background regions are largely suppressed in the attention maps. 
This behavior indicates that the action-query decoder learns to extract interaction-centric visual evidence from the current observation, supporting compact future state prediction in the learned latent space.

\subsection{Discussion on Alternative Architectural Designs}
To further validate the necessity of our architectural choices, we explored an alternative teacher-student training paradigm based on an Exponential Moving Average (EMA) momentum encoder, replacing the InfoNCE contrastive objective. 
Such EMA-based frameworks have been highly successful in self-supervised visual representation learning, as exemplified by methods such as BYOL~\citep{grill2020byol} and DINO~\citep{oquab2023dinov2, oquab2025dinov3}. 
However, in our action-conditioned dynamics modeling setting, we empirically found that this design quickly leads to severe representation collapse. 
We hypothesize that, without the explicit repulsion induced by negative samples in the contrastive objective, the action-conditioned predictor can exploit a trivial shortcut: it learns to ignore the action query and produces a nearly static representation that matches the slowly updated EMA target. 
This observation highlights the importance of hard negatives for enforcing action-dependent state transitions and preventing the learned future representation from collapsing to action-invariant features.

\section{Limitations}
\label{sec:app-e}

\paragraph{Limitations of contrastive supervision.}
Our Goal-Convergent Contrastive Objective learns action-conditioned transitions by aligning future predictions induced by different actions that converge to the same target.
While this design is effective for capturing action-conditioned visual transitions, its supervision signal is ultimately defined by the separability of transitions in the learned representation space. 
When multiple action sequences induce highly similar visual transitions, the contrastive signal may provide limited supervision for resolving fine-grained differences among these actions.
Consequently, CAPE can learn robust coarse-grained transition representations while being less sensitive to subtle action variations required for precise manipulation. 
This limitation may partly explain the performance gap on tasks that require accurate spatial alignment or fine-grained control. 
Future work could incorporate additional local consistency or geometric regularization to strengthen the model's sensitivity to small but task-critical action differences.

\paragraph{Limitations of full action-space modeling.}
Another limitation arises from the structure of the robot action space. 
As observed in JEPA-WM~\citep{terver2025jepa-wm}, the orientation and gripper dimensions in DROID are substantially harder to model than end-effector translation and can dominate action prediction errors due to noise, discontinuities, and contact-dependent dynamics. 
CAPE inherits this challenge: although it effectively models action-conditioned visual transitions, accurately predicting and optimizing over gripper and orientation actions remains difficult, especially in contact-rich manipulation scenarios such as grasping. 
Therefore, following JEPA-WM~\citep{terver2025jepa-wm}, our DROID offline Action Score evaluation focuses on the first three end-effector position dimensions. 
Future work could explore separate optimization procedures for gripper and orientation control to extend CAPE toward more precise contact-rich manipulation.


\newpage

\end{document}